\documentclass[10pt,twocolumn,letterpaper]{article}

\usepackage{cvpr}              

\usepackage{url}            
\usepackage{booktabs}       
\usepackage{amsfonts}       
\usepackage{nicefrac}       
\usepackage{microtype}      
\usepackage{xcolor}         

\usepackage{multirow}
\usepackage{graphicx}
\usepackage{subcaption}     
\usepackage{amsmath}
\usepackage{float}

\newlength{\mycolumnwidth}
\definecolor{cvprblue}{rgb}{0.21,0.49,0.74}
\usepackage[pagebackref,breaklinks,colorlinks,allcolors=cvprblue]{hyperref}

\def\paperID{17067} 
\def\confName{CVPR}
\def\confYear{2025}

\title{IML-ViT: Benchmarking Image Manipulation Localization by Vision Transformer}

\author{Xiaochen Ma\textsuperscript{1}, Bo Du\textsuperscript{1},  Zhuohang Jiang\textsuperscript{1}, Xia Du\textsuperscript{2}, Ahmed Y. Al Hammadi\textsuperscript{3},  and Jizhe Zhou\textsuperscript{1}  \\
\textsuperscript{1} Sichuan University \\
\textsuperscript{2} Xiamen University of Technology \\
\textsuperscript{3} Mohamed Bin Zayed University for Humanities
}

\begin{document}
\maketitle
\begin{abstract}
Advanced image tampering techniques are increasingly challenging the trustworthiness of multimedia, leading to the development of Image Manipulation Localization (IML). 
But what makes a good IML model? The answer lies in the way to capture artifacts. Exploiting artifacts requires the model to extract non-semantic discrepancies between manipulated and authentic regions, necessitating explicit comparisons between the two areas. With the self-attention mechanism, naturally, the Transformer should be a better candidate to capture artifacts. 
However, due to limited datasets, there is currently no pure ViT-based approach for IML to serve as a benchmark, and CNNs dominate the entire task. Nevertheless, CNNs suffer from weak long-range and non-semantic modeling.
To bridge this gap, based on the fact that artifacts are sensitive to image resolution, amplified under multi-scale features, and massive at the manipulation border, we formulate the answer to the former question as building a ViT with high-resolution capacity, multi-scale feature extraction capability, and manipulation edge supervision that could converge with a small amount of data. We term this simple but effective ViT paradigm IML-ViT, which has significant potential to become a new benchmark for IML. Extensive experiments on three different mainstream protocols verified our model outperforms the state-of-the-art manipulation localization methods. 
Code and models are available at \href{https://github.com/SunnyHaze/IML-ViT}{https://github.com/SunnyHaze/IML-ViT}.
\end{abstract}    
\section{Introduction}

\begin{figure}[t]
\centering
\includegraphics[width=0.95\columnwidth]{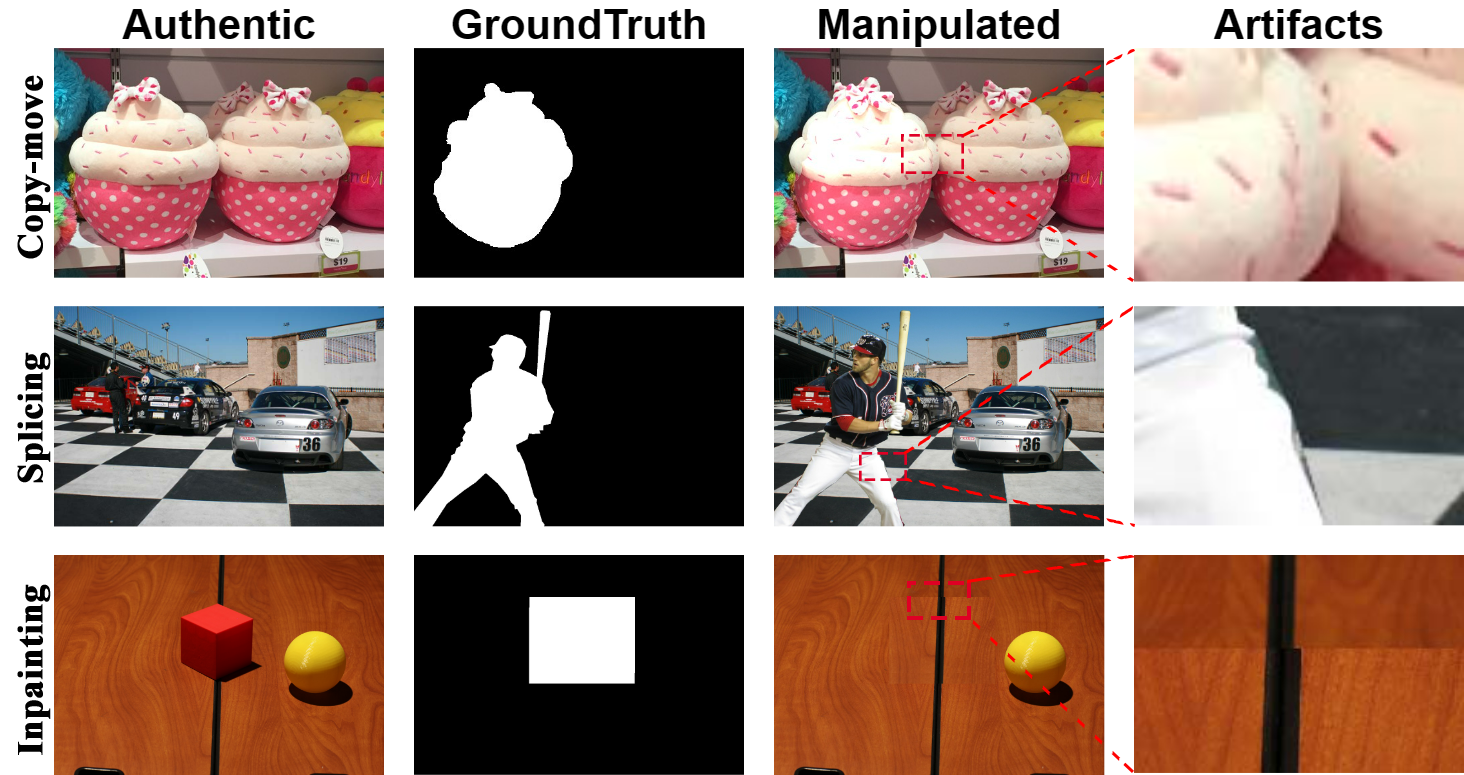}
\caption{
\textbf{An example of three types of manipulations and their corresponding artifacts.} Artifacts contain visible traces, including distortions, sudden changes, or anomalies caused by tampering operations. Artifacts are frequently found at the junction between two regions and appear in very detailed positions. For a better view, zooming in is recommended.
} 
\label{fig:artifact}
\end{figure}

With the advances in image editing technology like Photoshop, Image Manipulation Localization (IML) methods have become urgent countermeasures to cope with existing tampered images and avoid security threats~\cite{Verdoliva_2020}. Effective IML methods play a crucial role in discerning misinformation and have the potential to contribute to the safety of multimedia world. As shown in Figure \ref{fig:artifact}, the IML task aims to detect whether images have been modified and to localize the modified regions at the pixel level.
Image manipulation can be generally classified into three types~\cite{Mantra_2019, Verdoliva_2020}: (1) \textit{splicing}: copying a region from an image and pasting it to another image. (2) \textit{copy-move}: cloning a region within an image. (3) \textit{inpainting}: erasing regions from images and inpaint missing regions with visually plausible contents.


As shown in Table \ref{tab:1}, most existing methods for IML tasks greatly benefit from tracing artifacts with various CNN-based feature extractors. ``Artifacts" refer to unique visible traces (see Figure \ref{fig:artifact}) and invisible low-level feature inconsistencies (e.g., noise or high-frequency) resulting from manipulation.
As tampering aims to deceive the audience by creating semantically meaningful and perceptually convincing images, visual traces typically manifest at a non-semantic level, distributed in textures around the manipulated area.
Additionally, low-level features, like noise inconsistencies introduced by different cameras, can also serve as crucial evidence to reveal manipulated regions within the authentic area.
Thus, based on previous experiences, \textit{the key to IML lies in capturing the  artifacts by identifying non-semantic visible traces and low-level inconsistencies.}



\begin{table}[t]
\centering

\resizebox{\columnwidth}{!}{
\begin{tabular}{|c|cc|c|c|cc|}
\hline
\multirow{2}{*}{\textbf{Method}} &
  \multicolumn{2}{c|}{\textbf{Backbone }} &
  \multirow{2}{*}{\textbf{Resolution}} &
  
  \multirow{2}{*}{\textbf{Manipulation   supervision}} &
  \multicolumn{2}{c|}{\textbf{IML dataset thirsty}} \\ \cline{2-3} \cline{6-7} 
 &
  \multicolumn{1}{c|}{\textbf{CNN}} &
  \textbf{Tran.} &
   &
   &
  \multicolumn{1}{c|}{\textbf{Type}} &
  \textbf{Amount} \\ \hline
\begin{tabular}[c]{@{}c@{}}ManTra-Net\\      ~\cite{Mantra_2019}\end{tabular} &    

  \multicolumn{1}{c|}{\checkmark} &
  - &
  \begin{tabular}[c]{@{}c@{}}Resize\\  512×512\end{tabular}  &
  \begin{tabular}[c]{@{}c@{}}Noise\\      (BayarConv+ SRM filter)\end{tabular} &
  \multicolumn{1}{c|}{Private} &
  102k \\ \hline
\begin{tabular}[c]{@{}c@{}}SPAN\\      ~\cite{SPAN_2020}\end{tabular} &  
  \multicolumn{1}{c|}{\checkmark} &
  - &
   \begin{tabular}[c]{@{}c@{}}Resize\\  224×224\end{tabular}
 &
  \begin{tabular}[c]{@{}c@{}}Noise\\      (BayarConv+ SRM filter)\end{tabular} &
  \multicolumn{1}{c|}{Private} &
  102k \\ \hline
\begin{tabular}[c]{@{}c@{}}CR-CNN\\      ~\cite{CR_CNN_2020}\end{tabular} &  
  \multicolumn{1}{c|}{\checkmark} &
  - &
 \begin{tabular}[c]{@{}c@{}}Resize\\   short side to 600 \end{tabular}
  &
  \begin{tabular}[c]{@{}c@{}}Noise\\      (BayarConv+ SRM filter)\end{tabular} &
  \multicolumn{1}{c|}{Public} &
  5k \\ \hline
  \begin{tabular}[c]{@{}c@{}}GSR-Net\\      ~\cite{GSR_Net_2020}\end{tabular} &  
  \multicolumn{1}{c|}{\checkmark} &
  - &
   \begin{tabular}[c]{@{}c@{}}Resize\\   300×300 \end{tabular}
  &
  Edge Prediction &
  \multicolumn{1}{c|}{Public} &
  5k \\ \hline
   \begin{tabular}[c]{@{}c@{}}MVSS-Net\\      ~\cite{MVSS_2021}\end{tabular} & 
  \multicolumn{1}{c|}{\checkmark} &
  - &
     \begin{tabular}[c]{@{}c@{}}Resize\\   512×512 \end{tabular} &
  \begin{tabular}[c]{@{}c@{}}Noise (BayarConv)\\      Edge (sobel)\end{tabular} &
  \multicolumn{1}{c|}{Public} &
  5k \\ \hline
   \begin{tabular}[c]{@{}c@{}}MM-Net\\      ~\cite{mmnet_2021}\end{tabular} & 
  \multicolumn{1}{c|}{\checkmark} &
  - &
 \begin{tabular}[c]{@{}c@{}}Resize\\  short side to 800  \end{tabular}
  &
  Noise (BayarConv) &
  \multicolumn{1}{c|}{Private} &
  50k \\ \hline
    \begin{tabular}[c]{@{}c@{}}TransForensic\\    ~\cite{transforensic_2021}\end{tabular} & 

  \multicolumn{1}{c|}{\checkmark} &
  \checkmark &
\begin{tabular}[c]{@{}c@{}}Resize\\  512×512  \end{tabular}
  &
  - &
  \multicolumn{1}{c|}{Public} &
  10k \\ \hline
    \begin{tabular}[c]{@{}c@{}} ObjectFormer\\    ~\cite{objectformer_2022}\end{tabular} & 
  \multicolumn{1}{c|}{\checkmark} &
  \checkmark &
  \begin{tabular}[c]{@{}c@{}}Resize\\  256×256  \end{tabular}
  &
  High-frequency  &
  \multicolumn{1}{c|}{Private} &
  62K \\ \hline

    \begin{tabular}[c]{@{}c@{}}HiFi-Net \\    ~\cite{HiFi-Net2023}\end{tabular} & 
  \multicolumn{1}{c|}{\checkmark} &
  - &
\begin{tabular}[c]{@{}c@{}}Resize\\  256×256  \end{tabular}
&
  Frequency  &
  \multicolumn{1}{c|}{Public} &
  1,710K \\ \hline

    \begin{tabular}[c]{@{}c@{}}TruFor \\    ~\cite{trufor2023}\end{tabular} & 
  \multicolumn{1}{c|}{\checkmark} &
  \checkmark &
\begin{tabular}[c]{@{}c@{}}Crop\\ 512×512 \end{tabular}
  &
  \begin{tabular}[c]{@{}c@{}}Noise\\  (Contrastive learning)\end{tabular} &
  \multicolumn{1}{c|}{Public} &
  35K \\ \hline
  
\textit{
\begin{tabular}[c]{@{}c@{}}IML-ViT\\ (Ours) \end{tabular}
} &
  \multicolumn{1}{c|}{-} &
  \checkmark &
  \begin{tabular}[c]{@{}c@{}}  Zero-pad\\ 1024×1024 \end{tabular}
&
  Edge loss &
  \multicolumn{1}{c|}{Public} &
  5k \\ \hline
\end{tabular}
}
\vspace{2pt}
\caption{\textbf{Overview of State-of-the-Art End-to-End Models for Image Manipulation Localization.} \textit{Tran.} stands for \textit{Transformer}. \textit{Manipulation supervision} serves as prior knowledge broadly acknowledged in the image manipulation detection field. Edge information effectively traces visible artifacts, while noise and high-frequency features primarily highlight low-level differences between tampered and authentic regions.
}
\label{tab:1}
\end{table}

However, convolution propagates information in a \textit{collective} manner, making CNNs more suitable for semantic-related tasks, such as object detection, rather than tracing non-semantic artifacts that often surround an object. Further, to identify low-level inconsistencies, we need to explicitly compare the relationships between different regions. But in deeper networks, CNNs may overlook global dependencies~\cite{ERF_2016}, rendering them less effective in capturing differences between regions.
Given the weaknesses of CNN in non-semantic and long-distance modeling, we ask: \textit{Is there any other optimal backbone for solving IML tasks?} 

Considering the goal of capturing the feature discrepancies between the manipulated and authentic regions, we argue that self-attention should be a better solution regarding IML. \textit{As self-attention can explicitly model relationships between any areas regardless of their visual semantic relevance, especially for non-adjacent regions.} The performance boost achieved by SPAN~\cite{SPAN_2020} highlights the effectiveness of integrating self-attention structures into convolutional layers.
Furthermore, as artifacts are often distributed at the patch level rather than at the pixel or image level, Vision Transformer (ViT)~\cite{ViT_2021} naturally becomes the ideal choice to trace artifacts and make comparisons. 

While ViT may be suitable for IML tasks, directly applying the original ViT architecture is insufficient. We suggest that IML involves three key discrepancies from traditional segmentation tasks, which also have not yet received sufficient attention in previous IML methods, as supported by Table 1. These discrepancies are:

\textbf{High Resolution}
While semantic segmentation and IML share similar inputs and outputs, IML tasks are more information-intensive, focusing on detailed artifacts rather than macro-semantics at the object level. Existing methods use various extractors to trace artifacts, but their resizing methods already harm these first-hand artifacts.
Therefore, preserving the \textit{original resolution} of the images is crucial to retain essential artifacts for the model to learn.



\textbf{Edge Supervision} 
As mentioned earlier, IML's primary focus lies in detecting the distinction between the tampered and authentic regions. This distinction is most pronounced at the boundary of the tampered region, whereas typical semantic segmentation tasks only require identifying information within the target region. From another perspective, it becomes evident that visible artifacts are more concentrated along the periphery of the tampered region rather than within it (as shown in Figure \ref{fig:artifact}). Consequently, the IML task must guide the model to concentrate on the manipulated region's edges and learn its distribution for better performance.

\textbf{Multi-scale Supervision} The percentage of tampered area to the total area varies significantly across different IML datasets. CASIAv2~\cite{CASIA_2013} contains a considerable amount of sky replacement tampering, whereas Defacto~\cite{defacto_2019} mostly consists of small object manipulations. On average, CASIAv2 has 7.6\% of pixels as tampered areas, while Defacto has only 1.7\%. Additionally, IML datasets are labor-intensive and often limited in size, which poses challenges in bridging the gap between datasets.
Therefore, incorporating multi-scale supervision from the pre-processing and model design stages is essential to enhance generalization across different datasets.

\begin{figure*}
\centering
\includegraphics[width=1.4\columnwidth]{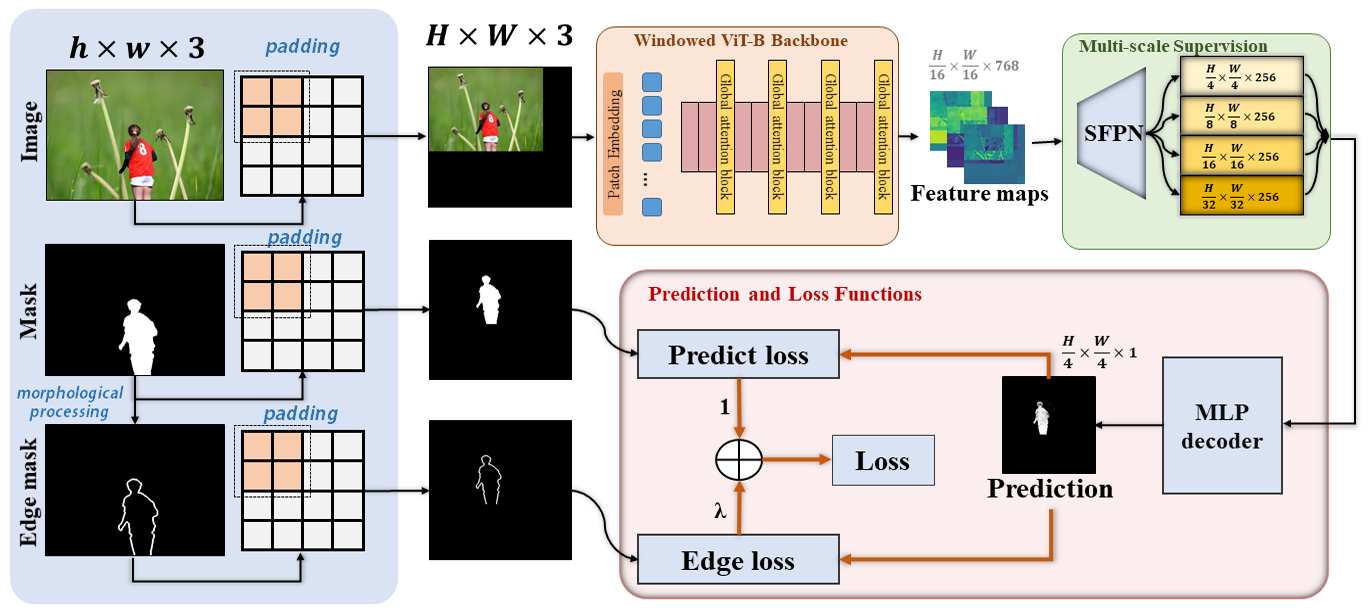}
\caption{\textbf{Overview of the general structure of IML-ViT. }  
} 
\label{fig:overall}
\end{figure*}

In this paper, we present IML-ViT, an end-to-end ViT-based model that solves IML tasks. Regarding the proposed three key discrepancies, we devise IML-ViT with the following components:
1) a windowed ViT which accepts \textbf{high-resolution} input. Most of the global attention block is replaced with windowed attention as the trade-off for time complexity. We initialize it with Masked Autoencoder (MAE)~\cite{MAE_2022} pre-trained parameters on ImageNet-1k~\cite{imagenet_2009}; 2) a simple feature pyramid networt (SFPN) ~\cite{benchmark_ViT_2021} to introduce \textbf{multi-scale supervision}; 3) a morphology-based edge loss strategy is proposed to ensure \textbf{edge supervision}. The overview of IML-ViT is shown in Figure \ref{fig:overall}.

In this manner, without any specialized modules, IML-ViT offers a general ViT structure for IML tasks. In other words, \textit{IML-ViT proves that IML tasks can be solved without hand-crafted features or deliberate feature fusion process}, promoting future IML methods into a more generalizable design paradigm. 

To the best of our knowledge, ObjectFormer~\cite{objectformer_2022}, TransForensics~\cite{transforensic_2021}, and TruFor~\cite{trufor2023} are the only Transformer-related models solving the IML tasks. However, their backbone distinguishes significantly from vanilla ViT, as will be explained in Section \ref{sec:related_works}. Thus, IML-ViT can be regarded as the pioneering model utilizing a vanilla ViT as the backbone for IML tasks.


Currently, the evaluation protocol for IML tasks is rather chaotic. To bring faithful evaluations and establish IML-ViT as the benchmark model, we demarcate existing messy evaluation settings into three mainstream protocols and conduct comprehensive experiments across these protocols. The extensive experiment results demonstrate that IML-ViT has surpassed all SoTA (state-of-the-art) models, thereby validating the reliability of the three proposed key essences of IML. Thus, we believe that IML-ViT is a powerful candidate to become a new SoTA model for IML.



In summary, our contributions are as follows:

\begin{itemize}
  \item We reveal the significant discrepancies between IML and traditional segmentation tasks by raising the three essences, which were overlooked by previous studies: high resolution, multi-scale, and edge supervision.
  
  \item  Aiming at three essences, we modify the components of ViT and establish the IML-ViT, the first ViT-based model for image manipulation localization.
  
  \item Extensive experiments show that IML-ViT outperforms state-of-the-art models in both $F_1$ and AUC scores on various protocols. This verifies the solidity of the three essences we proposed.
  
  \item We vanish the evaluation barrier for future studies by demarcating existing evaluation settings into three mainstream protocols and implementing cross-protocols-comparisons.
\end{itemize}

\section{Related Works}
\label{sec:related_works}
\par
\textbf{Paradigm of IML} 
Research in the early years focused on a single kind of manipulation detection, with studies on copy-move~\cite{Cozzolino_Poggi_Verdoliva_2015b, Rao_Ni_2016}, splicing~\cite{Cozzolino_Poggi_Verdoliva_2015a, Huh_Liu_Owens_Efros_2018, Kniaz_Knyaz_Remondino}, and removal (Inpainting)~\cite{Zhu_Qian_Zhao_Sun_Sun_2018}, respectively. However, since the specific type of tampering is unknown in practice, after 2018, general manipulation detection has become the focus. 
Many existing works follow the paradigm of ``feature extraction + backbone inference", especially extractors to exploit tamper-related information from artifacts. CR-CNN~\cite{CR_CNN_2020} has a noise-sensitive BayarConv~\cite{Bayar_2018} as the first convolution layer.  RGB-N networks~\cite{SRM_2018} develop an SRM filter to mine the difference in noise distribution to support decision-making.  ManTra-Net~\cite{Mantra_2019} and SPAN~\cite{SPAN_2020} combined SRM, BayarConv, and as the first layer of their model. Besides noise-related extractors, ObjectFormer \cite{objectformer_2022} employs a DCT module to extract high-frequency features, which are then combined with RGB features and fed into a transformer decoder. And MVSS-Net~\cite{MVSS_2021} combines a Sobel-supervised edge branch and a BayarConv noise branch with dual attention to fuse them. Nevertheless, a feature may only be effective for a single type of tampering, e.g., noise is more sensitive to splicing from different images but less effective for copy-move from the same image.  Recently, TruFor~\cite{trufor2023} and NCL~\cite{NCL_IML_2023} are the first to explore utilizing contrastive learning to extract features instead of manually designed filters. Proposed IML-ViT also aims to step out of the paradigm of ``extraction + fusion" and let the model itself learn as much knowledge as possible from the datasets rather than rely on \textit{priori knowledge}.

\par
\textbf{Transformer-based IML method}
At present, there are three Transformer-based models in the field of IML, namely ObjectFormer~\cite{objectformer_2022} TransForensics~\cite{transforensic_2021}, and TruFor~\cite{trufor2023}. 
Though named ``Trans" or ``Former", these models are hardly in line with vanilla ViT in overall structures and design philosophies. 
In particular, different from ViT directly embedding the patched images before encoding, the first two methods utilize several CNN layers to extract feature maps initially and subsequently employ Transformers for further encoding, leading to neglecting crucial first-hand low-level information. On the other hand, TruFor follows SegFormer~\cite{SegFormer_2021}'s encoder, using convolution layers instead of position embedding to integrate the position information for Transformer blocks, which overlooked key global dependencies to capture differences between regions. 

Moreover, in ObjectFormer's encoder, the ``query" inputs are learnable vectors representing object prototypes $o_i$, not image embeddings. As a result, it focuses on capturing dependencies between object prototypes and image tokens, whereas a standard ViT encoder solely models the relationship between image embeddings. Besides, ObjectFormer is pre-trained with a large tampering-oriented synthesized private dataset, while IML-ViT achieves better performance with pre-training on the more accessible ImageNet-1k dataset.

Further, TransForensics has a different way to apply Transformer blocks. While ViT uses these blocks sequentially, TransForensics employs them in parallel, wherein each feature map of an FCN output is decoded with a Transformer block, and then fused for the final output.

In short, IML-ViT can be considered the first IML method with a vanilla ViT as its backbone and could easily benefit from recently advanced algorithms related to ViT, proving that IML tasks do not require complex designs.



\section{Proposed Method}
In this section, we introduce our powerful IML-ViT paradigm, as shown in Figure \ref{fig:overall}, it consists of three main components: (1) a windowed ViT to balance the high-resolution inputs and the space complexity; (2) a \textit{simple feature pyramid network} (SFPN) to introduce multi-scale features; and (3) a lightweight MLP decoder head with additional edge supervision, which aids in focusing on artifact-related features and ensures stable convergence.

\subsection{ViT Backbone}

\textbf{High Resolution}
The ViT Encoder aims to mine the detailed artifacts and explore the differences between the suspicious areas. Thus, it is essential to preserve the \textbf{original} resolution of each image to avoid downsampling that could potentially distort the artifacts. However, when training in parallel, all images within a batch must have the same resolution. To reconcile these demands, we adopt a novel approach that has not been applied to any IML method before. Rather than simply rescaling images to the same size, we pad images and ground truth masks with zeros and place the image on the top-left side to match a larger constant resolution. This strategy maintains crucial low-level visual information of each image, allowing the model to explore better features instead of depending on handcrafted prior knowledge. To implement this approach, we first adjust the embedding dimensions of the ViT encoder to a larger scale. 

\textbf{Windowed Attention} To balance the computation cost from high resolution, we adopt a technique from previous works~\cite{MViTv2_2022, benchmark_ViT_2021}, which periodically replaces part of the global attention blocks in ViT with windowed attention blocks.  This method ensures global information propagation while reducing complexity. Differing from Swin~\cite{Swin_2021}, this windowed attention strategy is non-overlapping.

\textbf{MAE Pre-train} We initialize the ViT with parameters pre-trained on ImageNet-1k~\cite{imagenet_2009} with Masked Auto Encoder (MAE)~\cite{MAE_2022}. This self-supervised method can alleviate the over-fitting problem and helps the model generalize, supported by Table \ref{tab:ablation}.

More specifically, we represent input images as $X\in \mathbb{R}^{3\times h \times w }$, and ground truth masks as $M\in \mathbb{R}^{1\times h \times w }$, where $h$ and $w$ correspond to the height and width of the image, respectively. We then pad them to $X_p \in \mathbb{R}^{ 3 \times H \times W }$ and $M_p \in \mathbb{R}^{ 1 \times H \times W }$. Balance with computational cost and the resolution of datasets we employ in Table~\ref{tab:datasets}, we take $H=W=1024$ as constants in our implementation. Then $X_p$ is passed into the windowed ViT-Base encoder with 12 layers, with a complete global attention block retained every 3 layers. The above process can be formulated as follows:
\begin{equation}
G_e=\mathcal{V}(X_p) \in \mathbb{R}^{768 \times \frac{H}{16}\times \frac{W}{16} }
\end{equation}
where $\mathcal{V}$ denotes the ViT, and $G_e$ stands for encoded feature map. The number of channels, 768, is to keep the information density the same as the RGB image at the input, as $ 768 \times \frac{H}{16} \times  \frac{W}{16}  =  3 \times H \times W $.

\subsection{Simple Feature Pyramid Network}
To introduce multi-scale supervision, we adopt the \textit{simple feature pyramid} network (SFPN) after the ViT encoder, which was suggested in ViTDet~\cite{ViTDet_2022}. 
This method takes the single output feature map $G_e$ from ViT, and then uses a series of convolutional and deconvolutional layers to perform up-sampling and down-sampling to obtain multi-scale feature maps:
\begin{equation}
F_i = \mathcal{C}_i(G_e)  \in  \mathbb{R}^{ C_{S}\times \frac{H}{2^{i+2}}\times \frac{W}{2^{i+2}} } , i \in \{1,2,3,4\}
\end{equation}
Where $\mathcal{C}_i$ denotes the convolution series, and $C_S$ is the output channel dimension for each layer in SFPN. 
This multi-scale method does not change the base structure of ViT, which allowed us to easily introduce recently advanced algorithms to the backbone.

\subsection{Light-weight Predict Head}
\begin{figure}[t]
\centering
\includegraphics[width=1\columnwidth]{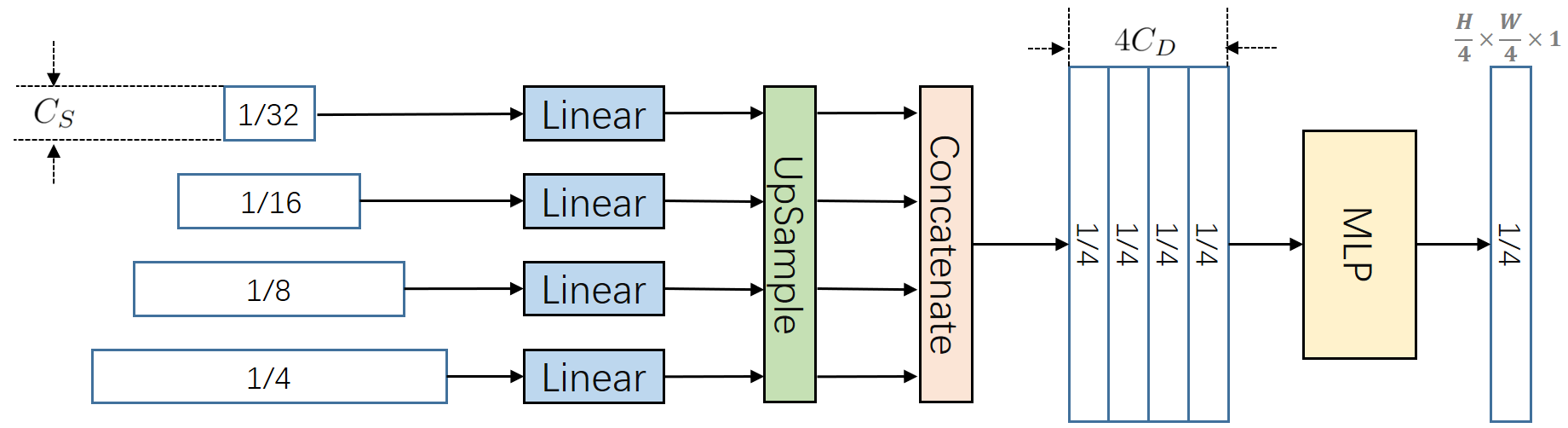}
\caption{\textbf{Diagrams of the predict-head.} The rectangles on the left represent the output of SFPN. There is a normalization layer before entering the MLP block, which is fully discussed below.} 
\label{fig:decoder}
\end{figure}

For the final prediction, we aimed to design a model that is simple enough to reduce memory consumption while also demonstrating that the improvements come from the advanced design in the ViT Encoder and the multi-scale supervision. Based on these ideas, we adopted the decoder design from SegFormer~\cite{SegFormer_2021}, which outputs a smaller predicted mask $M_e$ with a resolution of $1 \times \frac{H}{4}\times \frac{W}{4}$. The lightweight all-MLP decoder first applies a linear layer to unify the channel dimension. It then up-samples all the features to the same resolution of $  C_D \times \frac{H}{4}\times \frac{W}{4}$ with bilinear interpolation, and concatenates all the features together.
Finally, a series of linear layers is applied to fuse all the layers and make the final prediction. We can formulate the prediction head as follows:
\begin{equation}
P = \, MLP\{ \odot_i (W_iF_i+b_i) \}  \in \mathbb{R} ^{\frac{H}{4}\times \frac{W}{4} \times 1} 
\end{equation}
Here, $P$ represents the predicted probability map for the manipulated area; $\odot$ denotes concatenation operation, and $MLP$ refers to an MLP module. Detailed structure and analysis are illustrated in Figure \ref{fig:decoder}.

\subsection{Edge Supervision Loss}
\begin{figure}[t]
\centering
\begin{subfigure}{0.45\mycolumnwidth}
  \centering
  \includegraphics*[width=1.0\columnwidth]{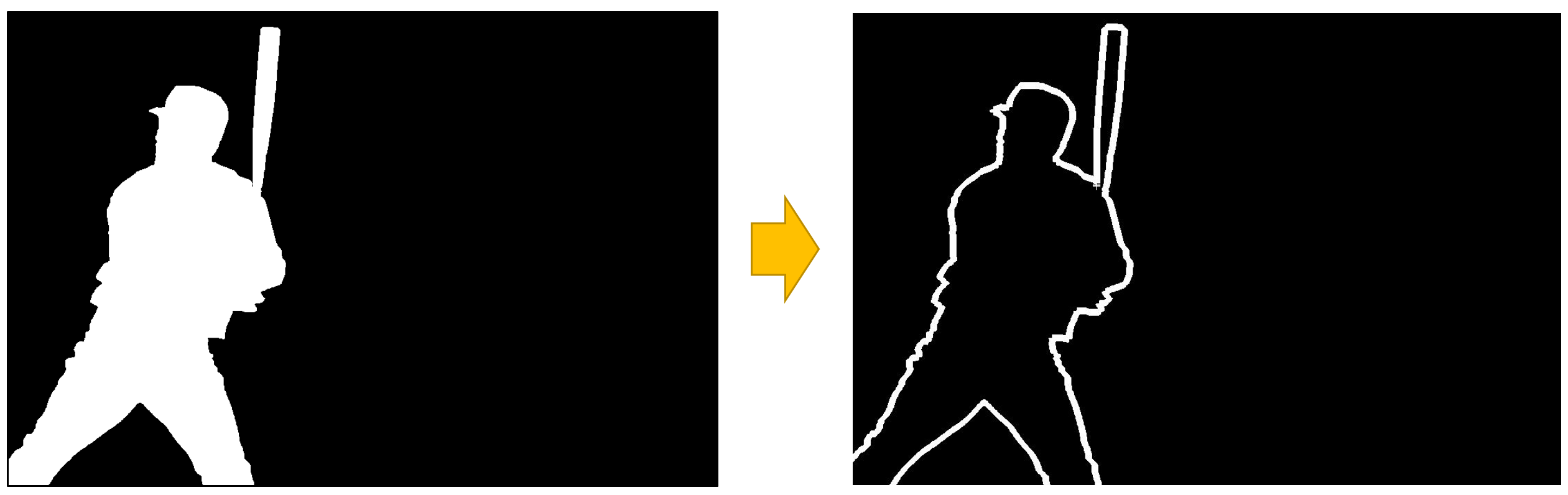}
  \caption{more manipulated region }
\end{subfigure}
\begin{subfigure}{0.45\mycolumnwidth}
  \centering
  \includegraphics*[width=1.0\columnwidth]{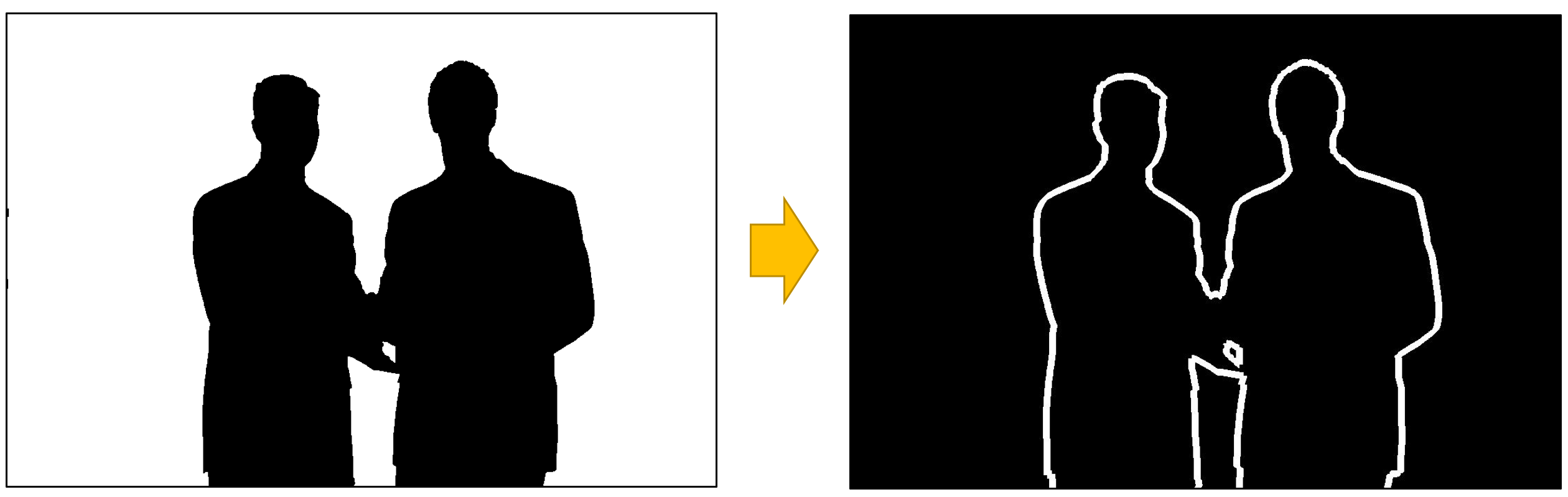}
  \caption{more authentic region}
\end{subfigure}
\caption{\textbf{Examples of generating the edge mask $M^\star$.} White region represents for manipulated area, $k$ is set to 7 while the image size is 1024×682.
The absolute value operation ensures that whether the tampered region dominates or the non-tampered region dominates, the mask only emphasizes the junction of the two.}
\label{fig:edge}
\end{figure}

To account for the fact that artifacts are typically more prevalent at the edges of tampered regions, where the differences between manipulated and authentic areas are most noticeable, we developed a strategy that places greater emphasis on the boundary region of the manipulated area. Specifically, we generate a binary edge mask $M^\star$ from the original mask image $M$ using mathematical morphology operations including dilation ($\oplus$) and erosion ($\ominus$) \cite{morphology_1983image}, followed by taking the absolute values of the result. The formula we use to generate the edge mask is:
\begin{equation}
M^\star = |(M \ominus B(k)) - (M \oplus B(k))|
\end{equation}
where, $B(x)$ generates a $(2x+1) \times (2x+1)$ \textit{cross} matrix, where only the $x^{th}$ column and $x^{th}$ row have a value of 1, while the rest of the matrix contains 0s. The integer value $x$ is selected to be approximately equal to the width of the white area in the boundary mask. Examples of the edge mask generated using this approach are shown in Figure \ref{fig:edge}.

\begin{figure}[t] \centering 
\includegraphics[width=1.0\columnwidth]{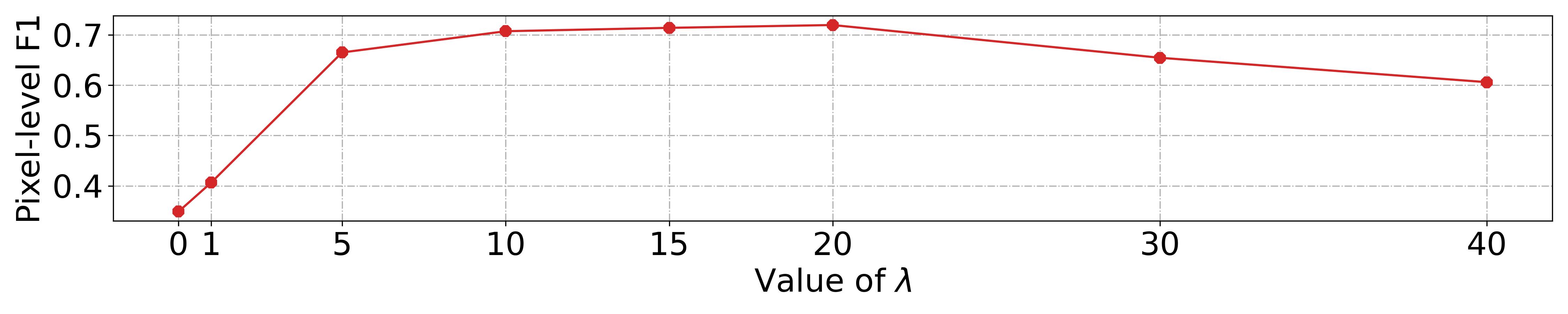} 
\caption{\textbf{Lambda selection, trained/test on CASIAv2/v1.}}  
\label{fig:lambda} 
\end{figure}

\textbf{Combined Loss}
To compute the loss function, we first pad the ground-truth mask $M$ and the edge mask $M^\star$ to the size of $H \times W$, and refer to them as $M_p$ and $M^\star_p$, respectively. We then calculate the final loss using the following formula:
\begin{equation}
\mathcal{L} = \mathcal{L}{seg}(P, M_p) + \lambda \cdot \mathcal{L}{edge}(P * M^\star_p, M_p * M^\star_p)
\end{equation}
where $*$ denotes the point-wise product, which masks the original image. Both $\mathcal{L}{seg}$ and $\mathcal{L}{edge}$ are binary cross-entropy loss functions, and $\lambda$ is a hyper-parameter that controls the balance between the segmentation and edge detection losses. By default, we searched the optimal $\lambda = 20$ to guide the model to focus on the edge regions, which is supported by Figure \ref{fig:lambda}. We choose a larger value for $\lambda$ also for two reasons: (1) to emphasize the boundary region, and (2) to balance the significant number of zeros introduced by zero-padding.

While the proposed edge loss strategy is straightforward, as we will discuss in the Experiments section (Figure \ref{fig:edge_loss}), it remarkably accelerates model convergence, stabilizes the training process, and mitigates potential NaN issues. Therefore, we consider this strategy a powerful prior knowledge for IML problems, deserving attention in future research.

\section{Experiments}


\subsection{Experimental Setup}

\textbf{Evaluation barrier for IML} 
While recent studies have introduced numerous SoTA models, comparing them on an equal footing remains challenging. This is due to the following reasons: 1) lack of publicly available code for the models and training processes~\cite{SPAN_2020, MVSS_2021}; 2) utilization of massive synthesized datasets that are inaccessible to the wider research community~\cite{Mantra_2019,mmnet_2021, objectformer_2022};
3) training and testing datasets often vary across different papers, also bringing difficulty for comparison.

\textbf{Datasets and Evaluation Protocol}
To facilitate reproducibility and overcome the existing evaluation barrier, we demarcate existing mainstream IML methods into three distinct protocols based on different partitions of datasets. Subsequently, we compare IML-ViT against SoTA methods with these three protocols, as shown in Table \ref{tab:datasets} and Table \ref{tab:protocols}. We followed MVSS-Net~\cite{MVSS_2021} to create Defacto-12k dataset. More details will be discussed in Section \ref{sec:compare_sota}.




\begin{table}[t]
\caption{\textbf{All datasets in our experiments.}}
\centering
\resizebox{\columnwidth}{!}{
\begin{tabular}{@{}cccccccc@{}}
\toprule[2pt]
\multirow{2}{*}{\textbf{Dataset}} & \multicolumn{2}{c}{\textbf{Type}}         & \multicolumn{3}{c}{\textbf{Manipulation type}}  & \multicolumn{2}{c}{\textbf{Resolution}} \\ \cmidrule(l){2-8} 
                                  & \textbf{Authentic} & \textbf{Manipulated} & \textbf{copymv} & \textbf{spli} & \textbf{inpa} & \textbf{min}       & \textbf{max}       \\ \midrule
CASIAv2~\cite{CASIA_2013}         & 7491               & 5123                 & 3274            & 1849          & 0             & 240                & 800                \\
CASIAv1~\cite{CASIA_2013}         & 800                & 920                  & 459             & 461           & 0             & 256                & 384                \\
NIST16~\cite{NIST16_2019}         & 0                  & 564                  & 68              & 288           & 208           & 480                & 5616               \\
COVERAGE~\cite{Coverage_2016}     & 100                & 100                  & 100             & 0             & 0             & 158                & 572                \\
Defacto-12k~\cite{defacto_2019}   & 6000               & 6000                 & 2000            & 2000          & 2000          & 120                & 640                \\
Columbia~\cite{Coverage_2016}     & 183                & 180                  & 0               & 180           & 0             & 568                & 1152               \\
IMD-20~\cite{IMD20_2020}          & 415                & 2010                 & -               & -             & -             & 176                & 4437               \\
tampCOCO~\cite{CAT-Net2022}       & 0                  & 800000               & 600000          & 200000        & 0             & 51                 & 640                \\
JPEG RAISE~\cite{CAT-Net2022}     & 24462              & 0                    & -               & -             & -             & 1515                  & 6159                  \\ \bottomrule[2pt]
\end{tabular}
}
\label{tab:datasets}
\end{table}

\begin{table}[t]
\centering
\caption{\textbf{The protocols we demarcate from the existing works.}}
\resizebox{\columnwidth}{!}{
\begin{tabular}{@{}c|l|l@{}}
\toprule[2pt]
\multicolumn{1}{c|}{\textbf{Protocol}} & \multicolumn{1}{c|}{\textbf{Details}}                          & \multicolumn{1}{c}{\textbf{Seminal paper}} \\ \midrule
\textit{No.1}                                   & Train on CASIAv2. Test on other small datasets.                & MVSS-Net~\cite{MVSS_2021}                                 \\

\textit{No.2}                                  & Train on six large mixed datasets. Test on other small datasets. & CAT-Netv2~\cite{CAT-Net2022}                                    \\ 
\textit{No.3}                                & Random split train/test dataset on mixed small datasets.       & ObjectFormer~\cite{objectformer_2022}                               \\

\bottomrule[2pt]
\end{tabular}
}

\label{tab:protocols}
\end{table}


\textbf{Evaluation Criteria} 
We evaluate our model using pixel-level $F_1$ score with a fixed threshold $0.5$ and Area Under the Curve (AUC), which are commonly used evaluation metrics in previous works. Both of them are metrics where higher values indicate better performance. However, it's worth noting that AUC can be influenced by excessive true-negative pixels in IML datasets, leading to an overestimation of model performance. Nevertheless, our model achieves SoTA performance in both $F_1$ score and AUC.

\textbf{Implementation}  We pad all images to a resolution of 1024x1024, except for those that exceed this limit. For the larger images, we resized them to the longer side to 1024 and maintained their aspect ratio. During training, following MVSS-Net~\cite{MVSS_2021}, common data augmentation techniques were applied, including re-scaling, flipping, blurring, rotation, and various naive manipulations (e.g., randomly copy-moving or inpainting rectangular areas within a single image).
We used the AdamW optimizer~\cite{AdamW_2019} with a base learning rate of 1e-4, scheduled with a cosine decay strategy~\cite{cosine_decay_2017}. The early stop technique was employed during training.

\textbf{Complexity}
Training IML-ViT with a batch size of 2 per GPU consumed 22GB of GPU memory per card. Using four NVIDIA 3090 GPUs, the model was trained on a dataset of 12,000 images over 200 epochs, taking approximately 12 hours. For inference, a batch size of 6 per GPU required 20GB of GPU memory, with an average prediction time of 0.094 seconds per image. Reducing the batch size to 1 decreased the GPU memory requirement to 5.4GB. We also compare the number of parameters and FLOPs with SoTA models in Table \ref{tab:compexity} and achieve highly competitive results.

\begin{table}[t]
\caption{\textbf{Complexity of IML-ViT compared to open-sourced SoTA models.} Inference time is measured on a per-image basis.}
\label{tab:compexity}
\centering
\resizebox*{1.0\columnwidth}{!}{ 
\begin{tabular}{@{}lccccc@{}}
\toprule[2pt]
\textbf{Method} & \textbf{Infer. Time(s)} & \textbf{Params.(M)} & \textbf{512×512 FLOPs(G)} & \textbf{1024×1024 FLOPs(G)} \\ \midrule
MVSS-Net\cite{MVSS_2021, mvsspp_2022}              & 2.929                            & 147                 & 167                       & 683                         \\
PSCC-Net~\cite{pscc_net_2022}               & 0.072                            & 3                   & 120                       & 416                         \\
HiFi-Net~\cite{HiFi-Net2023}                 & 1.512                            & 7                   & 404                       & 3470                        \\
TruFor~\cite{trufor2023}                & 1.231                            & 68                  & 231                       & 1016                        \\
IML-ViT                 & 0.094                            & 91                  & 136                       & 445                         \\ \bottomrule[2pt]
\end{tabular}
}
\end{table}

\begin{table*}[h]
\centering

\resizebox*{0.7\textwidth}{!}{ 
\begin{tabular}{@{}lcccccc@{}}

\toprule[2pt]
\multirow{2}{*}{\textbf{Method}} & \multicolumn{6}{c}{\textbf{Pixel-level $F_1$ score}}                                                               \\ \cmidrule(l){2-7} 
                               & \textbf{CASIAv1 
                               } & \textbf{Columbia
                               } & \textbf{NIST16
                               } & \textbf{Coverage
                               } & \textbf{Defacto-12k
                               } & \textbf{MEAN} \\ \midrule
HP-FCN*, ICCV19~\cite{HPFCN_2019}                 & 0.154   & 0.067    & 0.121 & 0.003    & 0.055       & 0.080 \\
ManTra-Net*, CVPR19~\cite{Mantra_2019}           & 0.155   & 0.364    & 0.000     & 0.286    & 0.155       & 0.192 \\
CR-CNN*, ICME20~\cite{CR_CNN_2020}       & 0.405   & 0.436    & 0.238 & 0.291    & 0.132       & 0.300 \\
GSR-Net*, AAAI20~\cite{GSR_Net_2020}      & 0.387   & 0.613    & 0.283 & 0.285    & 0.051       & 0.324 \\
MVSS-Net*, ICCV21~\cite{MVSS_2021}      & 0.452   & 0.638    & 0.292 & 0.453  & 0.137       & 0.394 \\
MVSS-Net (re-trained)& 0.435   & 0.303    & 0.203 & 0.329    & 0.097       & 0.270 \\
MVSS-Net++*, PAMI22~\cite{mvsspp_2022}  & 0.513 & 0.660  & 0.304 & \textbf{0.482} & 0.095 & 0.411 \\
NCL-IML, ICCV23~\cite{NCL_IML_2023} &0.598   & 0.704    &  0.231 & 0.383  & 0.066       & 0.396 \\
\textit{IML-ViT (ours)}                 & \textbf{0.721}   & \textbf{0.780}    & \textbf{0.331} & 0.410  & \textbf{0.156}   & \textbf{0.480} \\ \bottomrule[2pt]
\end{tabular}
}
\vspace{1pt}
\caption{\textbf{Evaluation results of Protocol No.1.} Except for ManTra-Net and HP-FCN, which were trained on a privately synthesized dataset, all the methods were trained on CASIAv2 datasets. The best scores are highlighted in bold. Symbol ``*" marks the results are quoted from MVSS-Net paper~\cite{MVSS_2021}. 
}
\label{tab:sota}
\end{table*}
\begin{table*}[t]
\centering
\caption{\textbf{Evaluation results of Protocol No.3.} $*$ marks cross-dataset results. Metrics are quoted.}
\resizebox{1.7\columnwidth}{!}{
\begin{tabular}{@{}l|l|cccc|cc@{}}
\toprule[2pt]
\multicolumn{1}{c|}{\multirow{2}{*}{\textbf{Method}}} &
  \multicolumn{1}{c|}{\multirow{2}{*}{\textbf{Datasets (Train/validate/test split)
  }}} &
  \multicolumn{4}{c|}{\textbf{Pixel-level AUC}} &
  \multicolumn{2}{c}{\textbf{Pixel-level $F1$}} \\ \cmidrule(l){3-8} 
\multicolumn{1}{c|}{} & \multicolumn{1}{c|}{}               & \textbf{COVER} & \textbf{NIST16\footnotemark}
& \textbf{CASIA} & \textbf{IMD-20
} & \textbf{CASIA} & \textbf{COVER} \\ \midrule
TransForesinc,
 ICCV21         & COVER + CASIA + IMD20 (8:1:1)           & 0.884          & -              & 0.850          & 0.848           & 0.627          & 0.674          \\
\textit{IML-ViT(Ours) }        & COVER + CASIA + IMD20 (8:1:1)          & 0.912          & 0.821* & \textbf{0.961} & \textbf{0.943}  & \textbf{0.825} & \textbf{0.815} \\ \midrule
ObjectFormer,
 CVPR22          & COVER(4:1); NIST(4:1); CASIA(v2:v1) &  0.957 & 0.996          & 0.882          & -               & 0.579          & 0.758    \\

Hifi-Net, CVPR23 &  COVER(4:1); NIST(4:1); CASIA(v2:v1) & \textbf{0.961} & 0.996 & 0.885 & - & 0.616 & 0.801  \\ \midrule

CFL-Net
, WACV23              & NIST16 + CASIA + IMD20 (8:1:1)          & -              & \textbf{0.997} & 0.863          & 0.899           & -              & -              \\
\textit{IML-ViT(Ours)  }       & NIST16 + CASIA + IMD20 (8:1:1)          & 0.801*        & \textbf{0.997} & \textbf{0.959} & \textbf{0.941}  & \textbf{0.820} & 0.505* \\ 
\bottomrule[2pt]
\end{tabular}
}

\label{tab:AUC}
\end{table*}

\begin{table}[h]
\centering

\resizebox*{0.8\columnwidth}{!}{ 
\begin{tabular}{@{}lll@{}}
\toprule[2pt]
\textbf{Method}      & \textbf{Pre-train}           & \textbf{$F_1$(\%)} \\ \midrule
RGB-N, CVPR18~\cite{RGBN_2018}        & ImageNet                             & 40.8                   \\
SPAN, ECCV20~\cite{SPAN_2020}        & Private synthesized dataset          & 38.2                   \\
Objectformer, CVPR22~\cite{objectformer_2022} & Private synthesized dataset          & 57.9                   \\
\textit{IML-ViT(Ours)}        & MAE on ImageNet-1k          & \textbf{72.0}                   \\ \bottomrule[2pt]
\end{tabular}
}
\caption{\textbf{Comparison with Closed-source methods with Protocol No.1.}  }
\label{tab:close}
\end{table}
\begin{table}[t]
\centering
\caption{\textbf{Pixel-level F1 on Protocol No.2.}}
\resizebox*{0.9\columnwidth}{!}{
\begin{tabular}{@{}llllll@{}}
\toprule[2pt]
Method                            & CASIAv1 & COVER & Columbia & NIST16 \\ \midrule
CAT-Netv2, IJCV22~\cite{CAT-Net2022} & 0.752   & 0.381 & 0.859    & 0.308  \\
TruFor, CVPR23~\cite{trufor2023}    & 0.737   & 0.600 & 0.859    & 0.399  \\
\textit{IML-ViT (ours)}           & \textbf{0.798}   & \textbf{0.654} & \textbf{0.945}    & \textbf{0.503}  \\ \bottomrule[2pt]
\end{tabular}
}
\label{tab:Trufor}
\end{table}

\begin{figure}[t]
  \centering
    \includegraphics[width=\columnwidth]{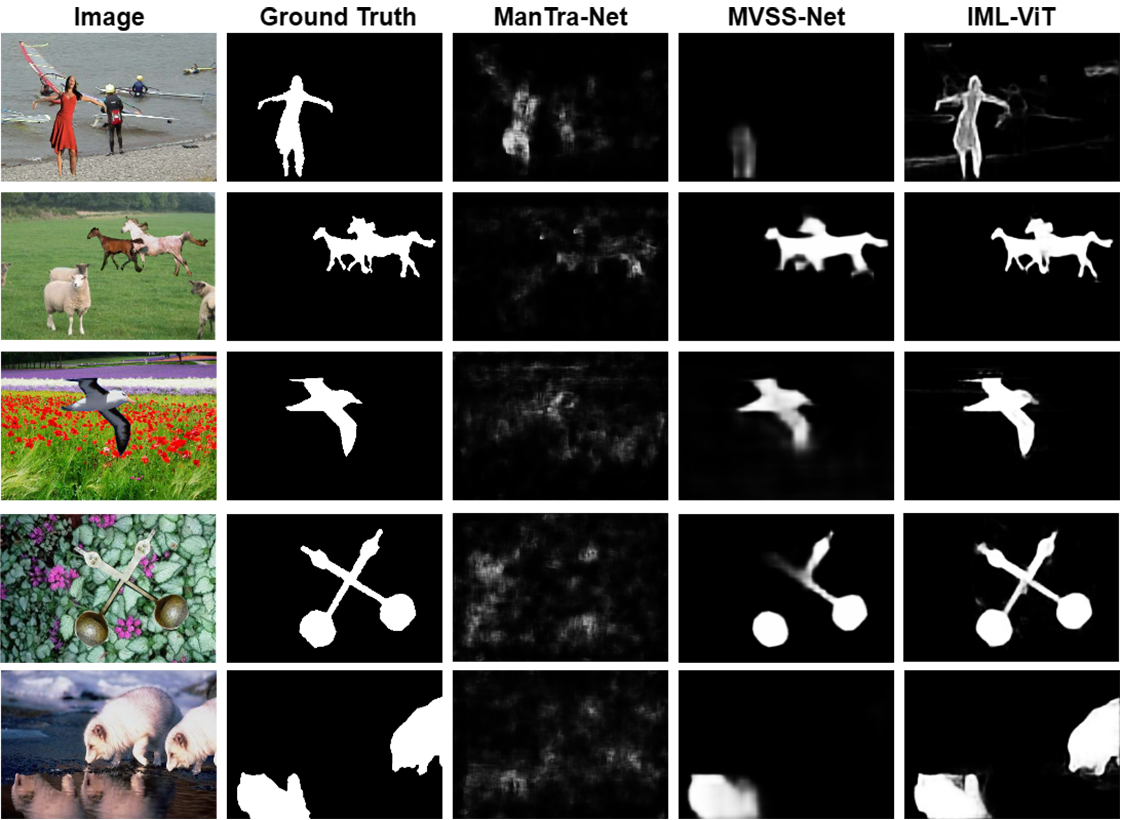}
  \caption{\textbf{Qualitative results on Protocol No.1 of IML-ViT compared to ManTra-Net and MVSS-Net. } For More results, see Appendix.
  }
  \label{fig:visualization}
\end{figure}

\subsection{Compare with SoTA (See Table \ref{tab:protocols} for protocols)}
\label{sec:compare_sota}

\textbf{Protocol No.1} 
Since MVSS-Net~\cite{MVSS_2021} has already conducted a detailed evaluation on a fair cross-dataset protocol and later works~\cite{NCL_IML_2023} followed their setting, we directly quote their results here and train our models with the same protocol.
The results measured by $F_1$ score are listed respectively in Table \ref{tab:sota}. 
We also compare this with some closed-source methods that only report their AUC tested on CASIAv1 in Table \ref{tab:close}.



Overall, our model achieves SoTA performance on this cross-dataset evaluation protocol. Figure \ref{fig:visualization}
illustrates that our model portrays high-quality and clear edges under different preferences of manipulation types.

\textbf{Protocol No.2} TruFor~\cite{trufor2023} is a recent strong method with extensive experimental results, training on six relatively large IML datasets proposed by CAT-Netv2~\cite{CAT-Net2022}. In our aim to establish IML-ViT as the benchmark model, we adopt their protocol to compare our model. We outperform them on four benchmark datasets. Details are shown in Table \ref{tab:Trufor}. 

\begin{figure}[t]
    \centering
    \includegraphics[width=\columnwidth]{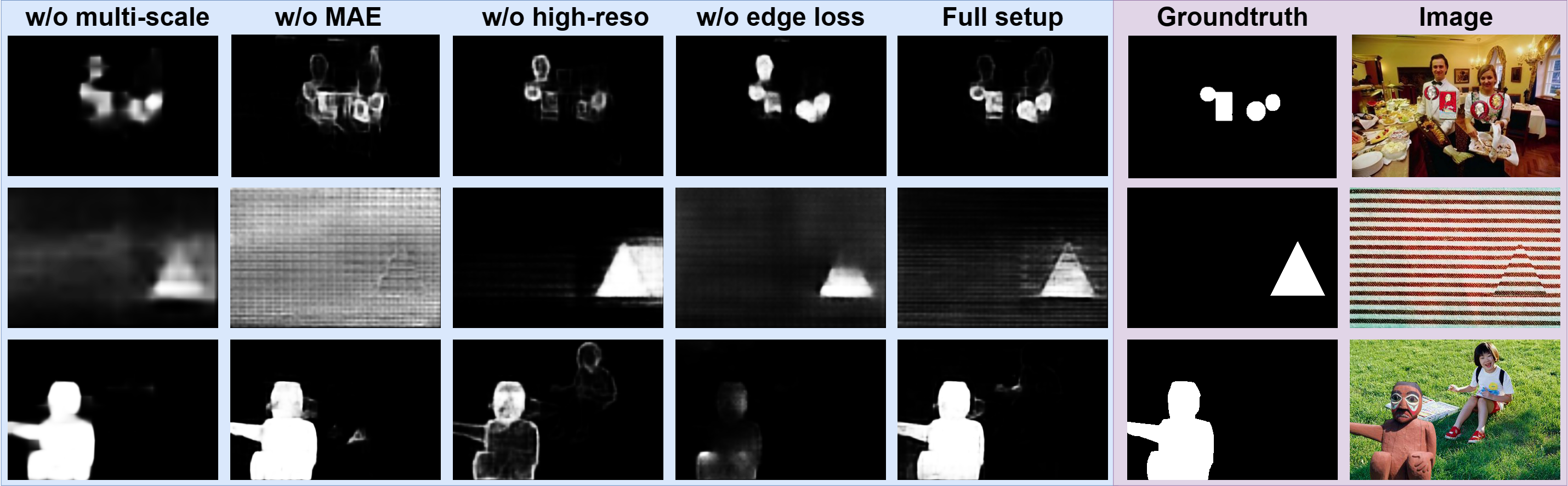}
    \caption{ Qualitative results of IML-ViT for ablation Study. We remove each component to test
their contribution.}
    \label{fig:ablation}
\end{figure}

\begin{figure}[t]
    \centering
\includegraphics[width=0.8\columnwidth]{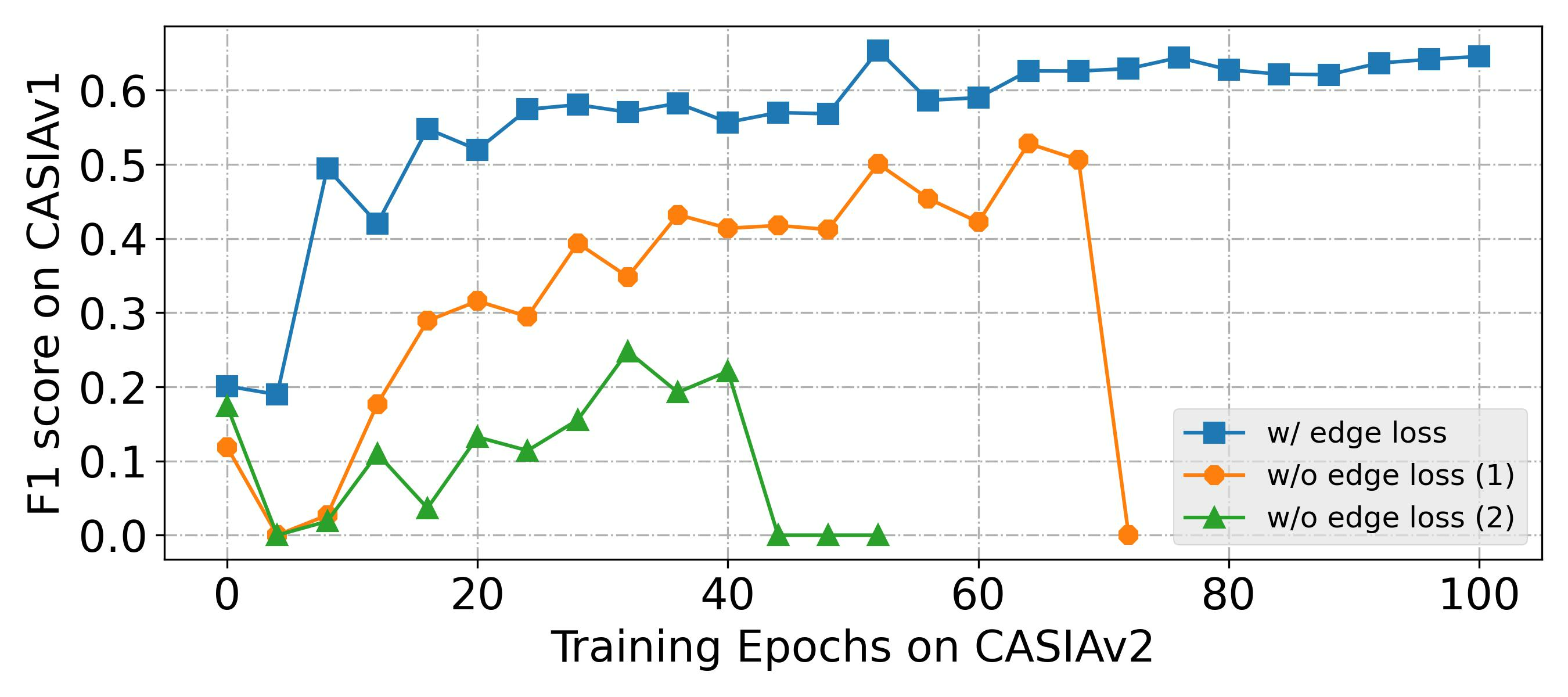}
    \caption{\textbf{Training stability influenced by proposed edge loss.}} 
    \label{fig:edge_loss}
\end{figure}

\begin{table*}[t]
\centering
\caption{\textbf{Ablation study of IML-ViT.} Each model is trained for 200 epochs on the CASIAv2 dataset. Best scores are marked in bold. 
\textit{H-Reso} refers to high resolution; \textit{SFPN} refers to simple feature pyramid network; and \textit{Edge} refers to proposed edge supervision.
}
\resizebox*{0.9\textwidth}{!}{ 
    \begin{tabular}{@{}c|c|ccc|cccccccccc@{}}

    \toprule[2pt]
    \multirow{2}{*}{\textbf{Test   Goal}} & \multirow{2}{*}{\textbf{Init Method}} & \multicolumn{3}{c|}{\textbf{Components}}        & \multicolumn{2}{c}{\textbf{CASIAv1}} & \multicolumn{2}{c}{\textbf{Coverage}} & \multicolumn{2}{c}{\textbf{Columbia}} & \multicolumn{2}{c}{\textbf{NIST16}} & \multicolumn{2}{c}{\textbf{MAEN}} \\ \cmidrule(l){3-15} 
                                          &                                       & \textbf{H-Reso} & \textbf{SFPN} & \textbf{Edge} & \textbf{F1}       & \textbf{AUC}     & \textbf{F1}       & \textbf{AUC}      & \textbf{F1}       & \textbf{AUC}      & \textbf{F1}      & \textbf{AUC}     & \textbf{F1}     & \textbf{AUC}    \\ \midrule
    \multirow{2}{*}{\textit{w/o MAE}}     & Xavier                                & +               & +             & +             & 0.1035            & -                & 0.0439            & -                 & 0.0744            & -                 & 0.0632           & -                & 0.0713          & -               \\
                                          & ViT-B ImNet-21k                       & +               & +             & +             & 0.5820            & 0.9037           & 0.2123            & 0.7898            & 0.5040            & 0.8335            & 0.2453           & 0.7939           & 0.3859          & 0.8302          \\
    \textit{w/o high resolution}          & MAE ImNet-1k                          & -               & +             & +             & 0.5747            & 0.9121           & 0.2622            & 0.7889            & 0.5150            & 0.8028            & 0.3292           & 0.7950           & 0.4153          & 0.8247          \\
    \textit{w/o multi-scale}              & MAE ImNet-1k                          & +               & -             & +             & 0.6504            & 0.9306           & 0.3877            & 0.8829            & 0.7096            & 0.8816            & 0.2847           & 0.7771           & 0.5081          & 0.8681          \\
    \textit{w/o edge-supervision}         & MAE ImNet-1k                          & +               & +             & -             & 0.6177            & 0.9240           & 0.3176            & 0.8789            & 0.6843            & 0.9161            & 0.2648           & 0.8045           & 0.4711          & 0.8809          \\
    \textit{Full setup}                   & MAE ImNet-1k                          & +               & +             & +             & \textbf{0.7206}   & \textbf{0.9420}  & \textbf{0.4099}   & \textbf{0.9137}   & \textbf{0.7798}   & \textbf{0.9337}   & \textbf{0.3317}  & \textbf{0.8064}  & \textbf{0.5605} & \textbf{0.8990} \\ \bottomrule[2pt]
    \end{tabular}

}
\vspace{2pt}

\label{tab:ablation}
\end{table*}

\begin{figure*}[h]
    \centering
    \includegraphics[width=0.85\textwidth]{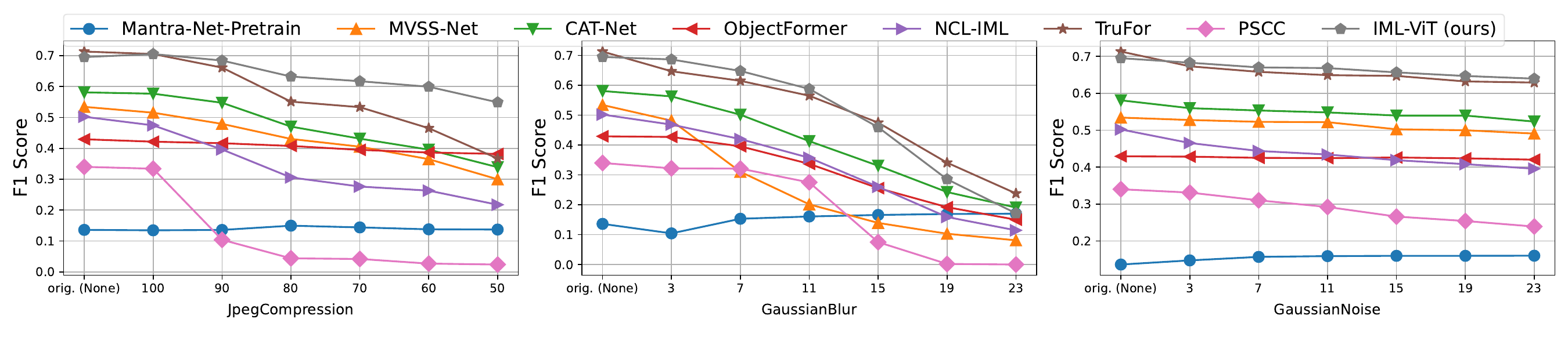}
    \caption{Robustness evaluation by three kinds of attacking across various SoTA methods under Protocol No.1.}
    \label{fig:robustness}
\end{figure*}

\textbf{Protocol No.3} 
TransForensic~\cite{transforensic_2021}, ObjectFormer~\cite{objectformer_2022}, HiFi-Net~\cite{HiFi-Net2023} and CFL-Net~\cite{CFL_Net_2023} reported their performance based on mixed datasets. 
They randomly split these datasets into training/validation/testing splits, causing the random splits performed by others to potentially differ, leading to a certain degree of unfairness.
Therefore, we do not recommend using this protocol in future work. However, for the sake of comparison with these state-of-the-art models, we also test IML-ViT following this protocol. 
Besides, note that HiFi-Net (1,710K images) and Objectformer (62k images) involve large IML datasets for pre-training, then tune on the specific small dataset, while we only pre-train with ImageNet-1k. Thus, we directly use results from mixed public IML datasets(14k images) to compare with them. Otherwise, it's easy to overfit on small datasets. In summary, experiment results in Table \ref{tab:AUC} show that, under this reasonable evaluation criteria, IML-ViT also outperforms these SoTA methods.



\subsection{Ablation Studies}
\label{lab:ablation}

To evaluate the contributions of each component to the model performance, we conducted experiments with multiple settings and compared them with a \textit{full setup} to test the four aspects we are most concerned about.
For \textit{initialization}, besides \textit{full setup} with MAE pre-training on ImageNet-1k, we test Xavier~\cite{Xavier_2010} initialization and ordinary ViT pre-training on ImageNet-21k.
To explore the impact of \textit{high resolution}, we resized all images to 512×512 during training before applying our padding strategy. For \textit{edge supervision}, we remove the edge loss for evaluation, while for \textit{multi-scale supervision}, we replace the module with the same number of plain convolution layers.

To reduce expenses, we trained model \textit{only} with \textbf{Protocol \textit{No.1}} in the ablation study. Qualitative results are illustrated in Figure \ref{fig:ablation}, which vividly demonstrates the efficacy of each component in our method. For quantitative results in Table \ref{tab:ablation}, our findings are:

\textbf{MAE pretrain is mandatory.} 
Indeed, dataset insufficiency is a significant challenge in building ViT-based IML methods. 
As shown in Table \ref{tab:datasets}, public datasets for IML are small in size, which cannot satisfy the appetite of vanilla ViT. As shown in \textit{w/o MAE}
aspects in Table \ref{tab:ablation}, the use of Xavier initialization to train the model resulted in complete non-convergence. However, while regular ViT pre-training initialization with Imagenet-21k achieves acceptable performance on CASIAv1, which is homologous to CASIAv2, it exhibits poor generalization ability on other non-homology datasets.
This indicates that MAE greatly alleviates the problem of non-convergence and over-fitting of ViT on limited IML datasets.

\textbf{Edge supervision is crucial.}
The performance of IML-ViT without edge loss shows significant variability with different random seeds, all leading to gradient collapse eventually, where the F1 score reaches 0, and the loss becomes \textit{NaN}, as shown in Figure \ref{fig:edge_loss}. In contrast, when employing edge loss, all performance plots exhibit consistent behavior similar to the blue line in Figure \ref{fig:edge_loss}, enabling fast convergence and smooth training up to 200 epochs. Furthermore, Table \ref{tab:ablation} confirms the effectiveness of edge loss in contributing to the final performance. In summary, these results demonstrate that edge supervision effectively stabilizes IML-ViT convergence and can serve as highly efficient prior knowledge for IML problems.

\textbf{High resolution is effective for artifacts.} 
The improved performance shown in Table \ref{tab:ablation} for the \textit{full setup} model across four datasets validates the effectiveness of the high-resolution strategy. However, it is essential to note that the NIST16 dataset shows limited improvement when using higher resolutions. This observation can be attributed to the fact that the NIST16 dataset contains numerous images with resolutions exceeding 2000, and down-sampling these images to 1024 for testing may lead to considerable distortion of the original artifacts, consequently reducing the effectiveness of learned features. Nevertheless, when considering the SoTA score achieved, it becomes evident that IML-ViT can flexibly infer the manipulated area based on the richness of different information types.

\textbf{Multi-scale supervision helps generalize.} All these datasets exhibit significant variations in the proportion of manipulated area, particularly where CASIAv2 has 8.96\% of the pixels manipulated, COVERAGE dataset has 11.26\%, Columbia dataset has 26.32\%, and NIST16 has 7.54\%. 
Nevertheless, the comprehensive improvements in Table \ref{tab:ablation} with the aid of multi-scale supervision indicate that this technique can effectively bridge the gap in dataset distribution, enhancing generalization performance.

\subsection{Robustness Evaluation}
We conducted a robustness evaluation on our IML-ViT model following MVSS-Net~\cite{MVSS_2021}. We utilized their protocol with three common types of attacks, including JPEG compression, Gaussian Noise, and Gaussian Blur. As shown in Figure \ref{fig:robustness}, IML-ViT achieved very competitive results among SoTA models, which proved to possess excellent robustness.

\section{Conclusions}
This paper introduces IML-ViT, the first image manipulation localization model based on ViT. Extensive experiments on three mainstream protocols demonstrate that IML-ViT achieves SoTA performance and generalization ability, validating the reliability of the three core elements of the IML task proposed in this study: high resolution, multi-scale, and edge supervision. Further, IML-ViT proves the effectiveness of self-attention in capturing non-semantic artifacts. Its simple structure makes it a promising benchmark for IML.

\newpage
{
    \small
    \bibliographystyle{ieeenat_fullname}
    \bibliography{main}
}

\newpage

\appendix

\section{Futher Robustness Evaluation}
\label{app:robustness}
\begin{figure}[h]
\centering
  \includegraphics[width=1.0\columnwidth]{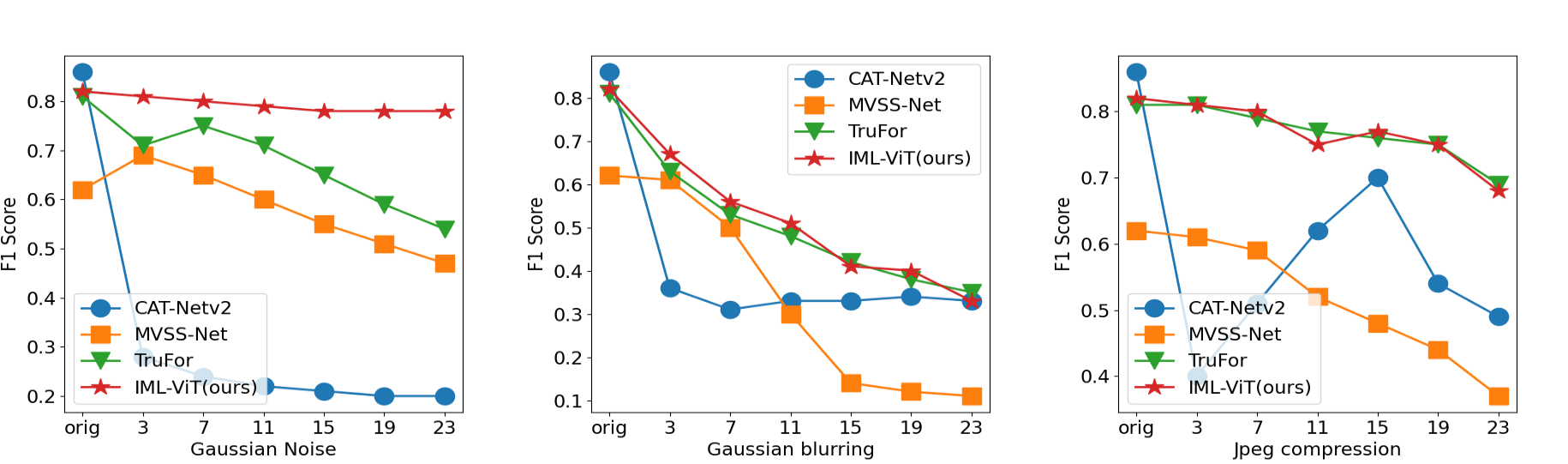}
  \caption{\textbf{Robustness evaluation against common attack on Protocol No.2.} Results are quoted from TruFor paper. They searched for the optimal F1 score to report the results while we selected 0.5 as the threshold for the F1 score here, proving IML-ViT more suitable for real-world scenarios.}
  \label{fig:robustness_trufor}
\end{figure}

\begin{figure}[h]
\centering
  \begin{subfigure}{0.45\columnwidth}
    \centering
    \includegraphics*[width=0.99\columnwidth]{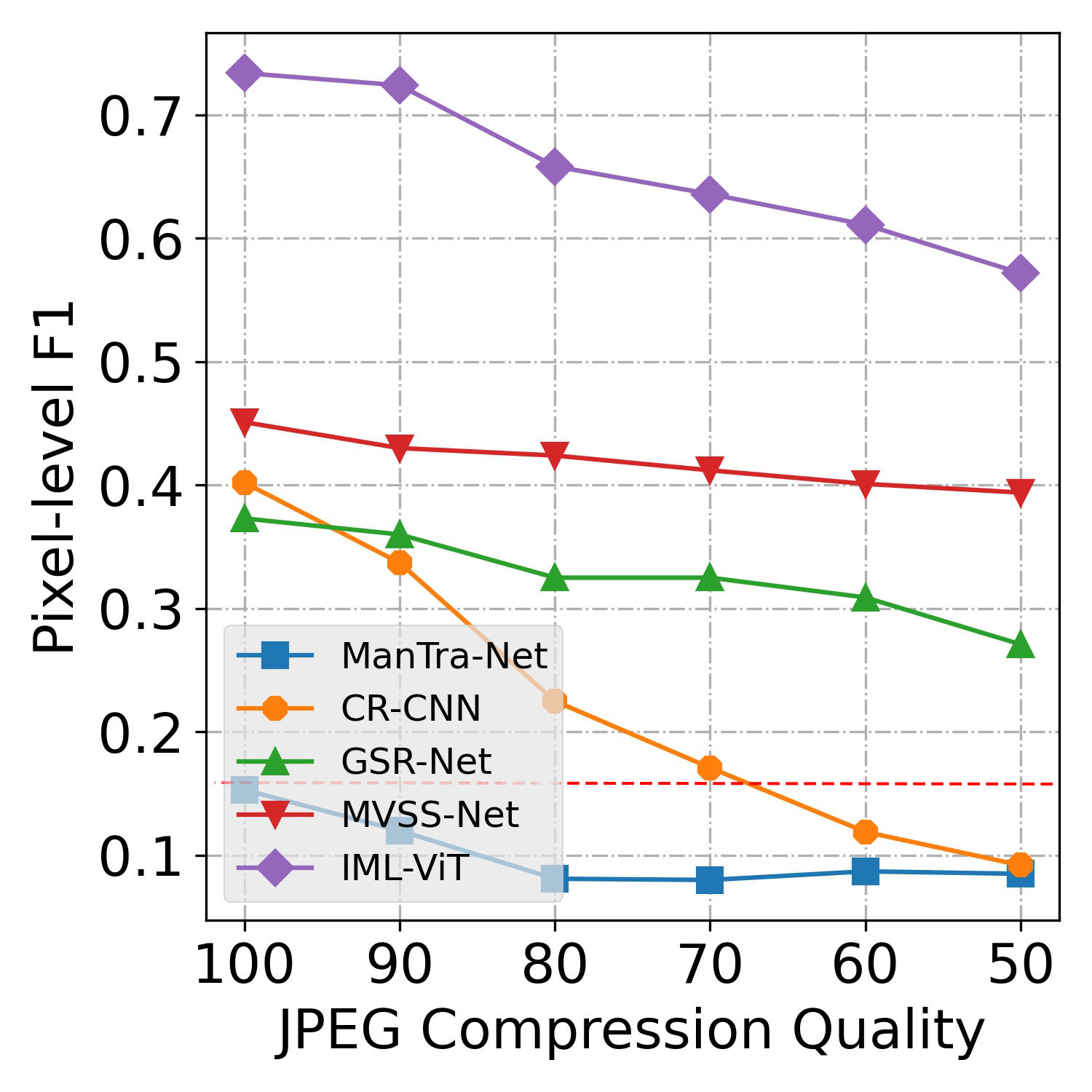}
  \end{subfigure}
  \begin{subfigure}{0.45\columnwidth}
    \centering
    \includegraphics*[width=0.99\columnwidth]{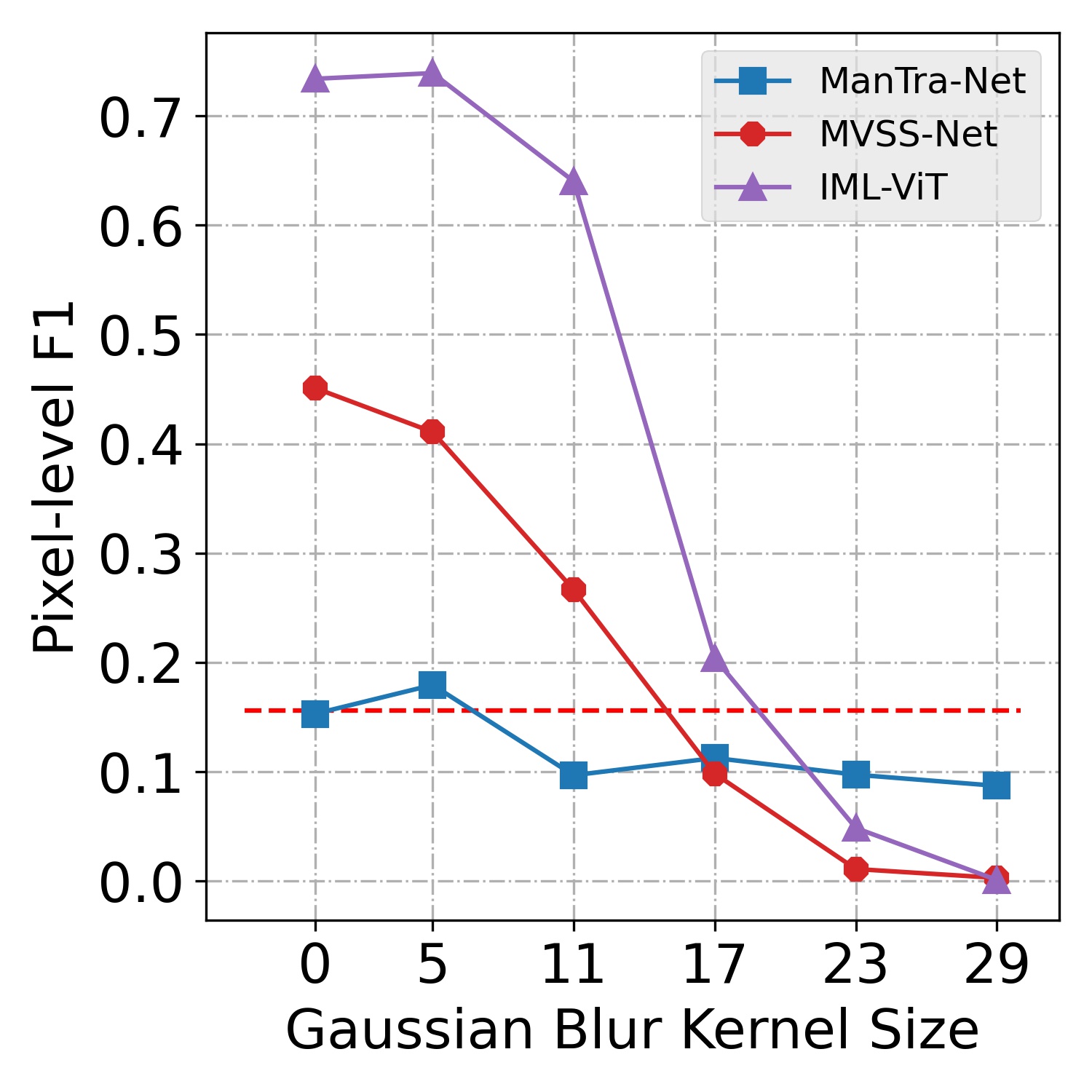}
  \end{subfigure}

  \caption{\textbf{Robustness Evaluation against JPEG compression and Gaussian blur on Protocol No.1.}
  The red dashed line represents the $F_1$ score when all predictions are classified as positive. 
  When the result is lower than this line, we consider the model to be less effective than random guessing and losing its localization ability.
  Performance against JPEG compression is quoted from MVSS-Net, while performance against Gaussian Blur is retested by us using three publicly available models.
  Our model has a later entry of the $F_1$ score into the red-line value compared to other models, and it consistently maintains a relatively high position, proving its better resistance.
  }
  \label{fig:robustness2}
\end{figure}

JPEG compression, Gaussian Noise, and Gaussian Blur are the common attack methods for Image manipulation localization. Following the convention from TruFor~\cite{trufor2023} and MVSS-Net~\cite{MVSS_2021}, we further carried out experiments on the resistance of these operations on Protocol No.1 and No.2. The evaluation results are shown in Figure \ref{fig:robustness_trufor} and  Figure \ref{fig:robustness2}. The IML-ViT exhibited excellent resistance to these attack methods and consistently maintained the best performance of the models.

\section{More Implementation Details}
\label{app:implementation details}

\subsection{High-resolution ViT-Base} Mostly following the original Vision Transformer, we implemented our model with a stack of Transformer blocks stacking together. LN are employed in the self-attention head and MLP blocks. Every two windowed attention blocks are followed by a global attention block. The windowed attention block only computes self-attention in a small, non-overlapped window, while the global attention block ensures global information propagation. Although we introduce the windowed attention, it only affects the self-attention manner but doesn't change the linear projection for Q, K, and V. Therefore, we can directly apply the MAE pre-trained parameters from a vanilla ViT-B with all global attention to this windowed ViT-B without any extra process. Detailed configuration are shown in Table \ref{tab:vitb}.

\begin{table}[h]
   \centering
   \resizebox{0.6\columnwidth}{!}{
   \begin{tabular}{@{}l|l@{}}
   \toprule[2pt]
   Configs              & Value     \\ \midrule
   patch size           & 16        \\
   embedding dim        & 768       \\
   depth                & 12        \\
   number of heads      & 12        \\
   input size           & 3×1024×1024 \\
   window size          & 14        \\
   Norm layer           & LN        \\
   Global block indexes & 2,5,8,11  \\ 
   Output shape & 768×64×64         \\ \bottomrule[2pt]
   \end{tabular}
   }
   \vspace{2pt}
   \caption{\textbf{Detailed structure of windowed ViT-Base}}
   \label{tab:vitb}
\end{table}

\subsection{Simple Feature Pyramid} After obtaining the output from ViT-B, SFPN utilizes a sequence of convolutional, pooling, or deconvolutional (ConvTranspose2D) layers to downsample it into feature maps with 256 channels, scaling them to resolutions of $\{4.0, 2.0, 1.0, 0.5, 0.25\}$ relative to the resolution of the input feature maps (768×64×64). For example, the largest output feature map with a scale of 4.0 is shaped like 256×256×256, while the smallest one with a scale of 0.25 is shaped like 256×8×8. Each layer is followed by LayerNorm. Detailed structures of each scale can be seen in Table \ref{tab:sfpn}.

\begin{table}[h]
   \centering
   \resizebox{0.9\columnwidth}{!}{
   \begin{tabular}{l|l}
   \toprule[2pt]
   Scales & Layers \& channels of feature maps     \\ \midrule
   4.0  & 768 ConvT 384 ConvT 192 Conv(1,1) 256 Conv(3,3) 256 \\
   2.0  &  768 ConvT 384 Conv(1,1) 256 Conv(3,3) 256 \\
   1.0  & 768 Conv(1,1) 256 Conv(3,3) 256 \\
   0.5  & 768 maxpool2D 384 Conv(1,1) 256 Conv(3,3) 256 \\
   0.25  & 768 maxpool2D 384 Conv(1,1) 256 Conv(3,3) 256 maxpool2D 256 \\ \bottomrule[2pt]
   \end{tabular}
   }
   \vspace{2pt}
   \caption{\textbf{Detailed structure of Simple feature pyramid} \textit{ConvT} denotes for ConvTranspose2D with kernel size of 2 and stride of 2; Conv(x,x) indicate that a Conv2D layer with kernel size of x; and maxpool2D has also a kernel size of 2. The number shown between layers indicates the number of channels for its respective feature map between layers. }
   \label{tab:sfpn}
\end{table}



\subsection{Predict-head's norm \& training including authentic images}
\label{sec:predict_heads}


The exact structure we applied in the predict-head is shown in Figure \ref{fig:decoder}. There is a norm layer before the last 1 × 1 convolution layer in the predict-head. We observed that when changing this layer may influence the following aspects: 1) convergence speed, 2) performance, and 3) generalizability. 

In particular, Layer Norm can converge rapidly but is less efficient at generalization. Meanwhile, the Batch Norm can be generalized better on other datasets. However, when including authentic images during training, the Batch Norm may sometimes fail to converge. At present, a straightforward solution is to use Instance Normalization instead, which ensures certain convergence.  Our experimental results are shown in Table \ref{tab:norms}.

Delving into the reasons, MVSS-Net~\cite{MVSS_2021} is the pioneering paper proposing the incorporation of authentic images with fully black masks during training to reduce \textit{false positives}.  We highly endorse this conclusion as it aligns more closely with the practical scenario of filtering manipulated images from a massive dataset of real-world images. However, in terms of metrics in convention, because the F1 score is meaningful only for manipulated images (as there are no positive pixels for fully black authentic images,
$F1 = \frac{2 \cdot \text{TP}}{2 \cdot \text{TP} + \text{FP} + \text{FN}}$, yielding $F1=0$), we only computed data for manipulated images. This approach may result in an ``unwarranted" metric decrease when real images are included.

\begin{table}[h]
   \centering
   \resizebox*{1.0\columnwidth}{!}{
\begin{tabular}{@{}c|c|cc|cc|cc|cc|c@{}}
\toprule[2pt]
\multirow{2}{*}{\textbf{Norm}} & \multirow{2}{*}{\textbf{Dataset}} & \multicolumn{2}{c|}{\textbf{CASIAv1}} & \multicolumn{2}{c|}{\textbf{Coverage}} & \multicolumn{2}{c|}{\textbf{Columbia}} & \multicolumn{2}{c|}{\textbf{NIST16}} & \multirow{2}{*}{\textbf{MEAN}} \\ \cmidrule(lr){3-10}
                               &                                   & \textbf{F1}      & \textbf{Epoch}     & \textbf{F1}      & \textbf{Epoch}      & \textbf{F1}      & \textbf{Epoch}      & \textbf{F1}     & \textbf{Epoch}     &                                \\ \midrule
Layer                          & CASIAv2-5k                        & 0.686           & 168                & 0.347           & 168                 & 0.760           & 128                 & 0.231          & 104                & 0.506                        \\
Batch                          & CASIAv2-5k                        & 0.702           & 184                & 0.421           & 184                 & 0.730           & 184                 & 0.317          & 184                & 0.543                        \\
Instance                       & CASIAv2-5k                        & 0.719           & 184                & 0.419           & 176                 & 0.792           & 140                 & 0.263          & 150                & 0.547                       \\
Batch                          & CASIAv2-12k                       & 0.715           & 176                & 0.352           & 128                 & 0.767           & 150                 & 0.263          & 124                & 0.524                        \\
Instance                       & CASIAv2-12k                       & 0.721           & 140                & 0.362           & 100                 & 0.784           & 136                 & 0.258          & 68                 & 0.531                         \\ \bottomrule[2pt]
\end{tabular}
   }
   \vspace{2pt}
   \caption{\textbf{Testing for norm layer in predict-head} Implementation is followed to ablation study in the main paper. CASIAv2-5k refers to manipulated images only, while 12k includes authentic images as well.}
   \label{tab:norms}
\end{table}

\textbf{Training settings}
Since our model could only train with small batch size, we applied the \textit{gradient accumulate} method during training, i.e. updating the parameters every 8 images during training. We select this parameter by experiments, details see Table \ref{tab:accumulate_grad}.

\begin{table}[h]
\centering
\resizebox*{1.0\columnwidth}{!}{
\begin{tabular}{@{}c|c|c|cc|cc|cc|cc|c@{}}
\toprule[2pt]
\multirow{2}{*}{\textbf{Batchsize}} & \multirow{2}{*}{\textbf{GPUs}} & \multirow{2}{*}{\textbf{accum iter}} & \multicolumn{2}{c|}{\textbf{CASIAv1}} & \multicolumn{2}{c|}{\textbf{Coverage}} & \multicolumn{2}{c|}{\textbf{Columbia}} & \multicolumn{2}{c|}{\textbf{NIST16}} & \multirow{2}{*}{\textbf{MEAN}} \\ \cmidrule(lr){4-11}
                                    &                                &                                      & \textbf{F1}      & \textbf{Epoch}     & \textbf{F1}      & \textbf{Epoch}      & \textbf{F1}      & \textbf{Epoch}      & \textbf{F1}     & \textbf{Epoch}     &                                \\ \midrule
2                                   & 4                              & 2                                    & 0.686            & 184                & 0.302            & 144                 & 0.685            & 92                  & 0.304           & 144                & 0.494                          \\
2                                   & 4                              & 4                                    & 0.704            & 192                & 0.386            & 140                 & 0.772            & 60                  & 0.331           & 140                & 0.548                          \\
2                                   & 4                              & 8                                    & 0.722            & 152                & 0.410            & 140                 & 0.780            & 84                  & 0.332           & 140                & 0.561                          \\
2                                   & 4                              & 16                                   & 0.706            & 184                & 0.419            & 184                 & 0.782            & 92                  & 0.314           & 184                & 0.555                          \\
2                                   & 4                              & 32                                   & 0.602            & 184                & 0.249            & 184                 & 0.740            & 184                 & 0.254           & 184                & 0.461                          \\ \bottomrule[2pt]
\end{tabular}
}
\vspace{2pt}
\caption{\textbf{Test for best accumulate gradient parameter for IML-ViT.} Train/tested on CASIAv2/v1 with four NVIDIA 3090 GPUs.}
\label{tab:accumulate_grad}
\end{table}

Besides, we adopt the early stop method during training. Evaluate the performance on the F1 score for CASIAv1, and stop training when there is no improvement for 15 epochs. Other configs are described in Table \ref{tab:vitb_train}.

\begin{table}[h]
   \centering
  \caption{\textbf{Training settings for IML-ViT}}
   \resizebox{0.8\columnwidth}{!}{
   \begin{tabular}{@{}l|l@{}}
   \toprule[2pt]
   Configs              & Value     \\ \midrule
   batch size & 2 (RTX 3090) or 4 (A40) \\
   GPU numbers & 4 (RTX 3090) or 2 (A40) \\
   accumulate gradient batch size & 8 \\
   epochs        & 200      \\
   warm up epochs & 4        \\
   optimizer & AdamW \\
   optimizer momentum & $\beta_1,\beta_2=0.9,0.95$ \\
   base  learning rate & 1e-4 \\
   minimum laerning rate & 5e-7 \\
   learning rate schedule & cosine decay \\
   weight decay  & 0.05 \\ 
   \bottomrule[2pt]
   \end{tabular}
   }
   \label{tab:vitb_train}
\end{table}

\section{What artifacts does IML-ViT capture?}
\label{sec:gradcam}
To investigate whether IML-ViT focuses on subtle artifacts as expected, we employ GradCAM~\cite{grad_cam_2017} to visualize the regions and content the model focuses on, as shown in Figure \ref{fig:grad_cam}. Additional results are in the Appendix \ref{sec:extra_gradcam}. We can observe that IML-ViT captures the traces around the manipulated region with the help of edge loss. Further, we can observe some extra subtle attention out of the manipulated region in the fourth image, proving the global dependent ability of ViT can help the model trace the tampered region.

\begin{figure}[h]
    \centering
    \includegraphics[width=1.0\columnwidth]{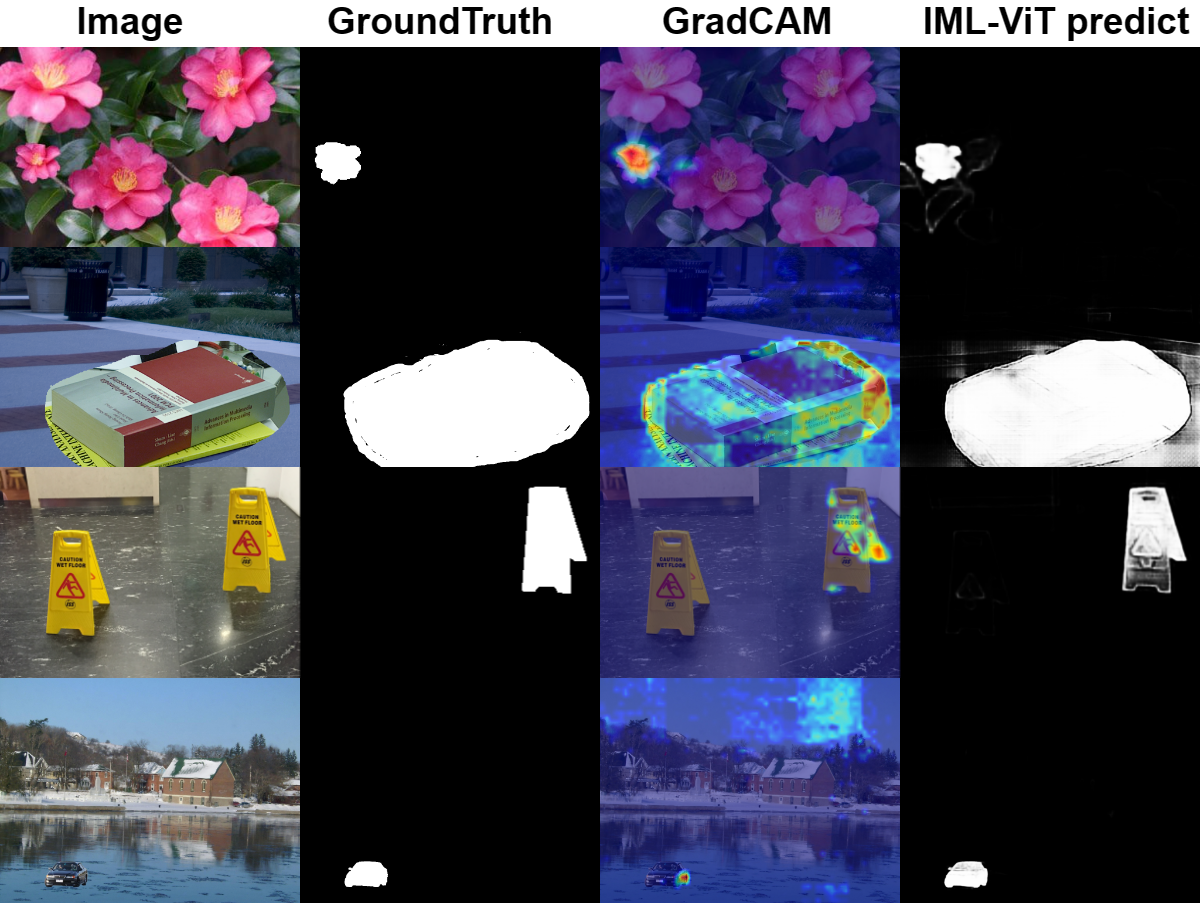}
    \caption{\textbf{GradCAM visualization of IML-ViT.}} 
    \label{fig:grad_cam}
\end{figure}

\section{Additional Experiments Results}
\label{app:additional_results}
Since the space limit, we place a part of our results in \textbf{Protocol No.1} here.




ObjectFormer~\cite{objectformer_2022} and CFL-Net~\cite{CFL_Net_2023} evaluate their models fine-tuning with CASIAv2 on AUC. Although this metric may overestimate the models, IML-ViT has still surpassed them, as shown in Table \ref{tab:auc_casia}.

\begin{table}[h]
\centering
\resizebox{0.9\columnwidth}{!}{
\begin{tabular}{@{}llllll@{}}
\toprule[2pt]
\textbf{Method}                & \textbf{CASIAv1}  & \textbf{Coverage} & \textbf{Columbia} & \textbf{NIST16} & \textbf{MEAN}     \\ \midrule

ObjectFormer~\cite{objectformer_2022} & 0.882 &- &- &-&- \\
CFL-Net~\cite{CFL_Net_2023} & 0.863 & - & - & 0.799 & - \\
\textit{IML-ViT(Ours)} & \textbf{0.931} & \textbf{0.918} & \textbf{0.962} & \textbf{0.818} & \textbf{0.917} \\
\bottomrule[2pt]
\end{tabular}
}
\caption{\textbf{Comparison of AUC scores trained on CASIAv2.}}
\label{tab:auc_casia}
\end{table}

\section{Extra Visualization} 

\subsection{Visualization of Protocol No.1 on other datasets}
\label{app:visualization_protocol_1}
\begin{figure}[t]
  \centering
  \begin{subfigure}{0.45\textwidth}
    \centering
    \includegraphics*
    [width=0.9\columnwidth]{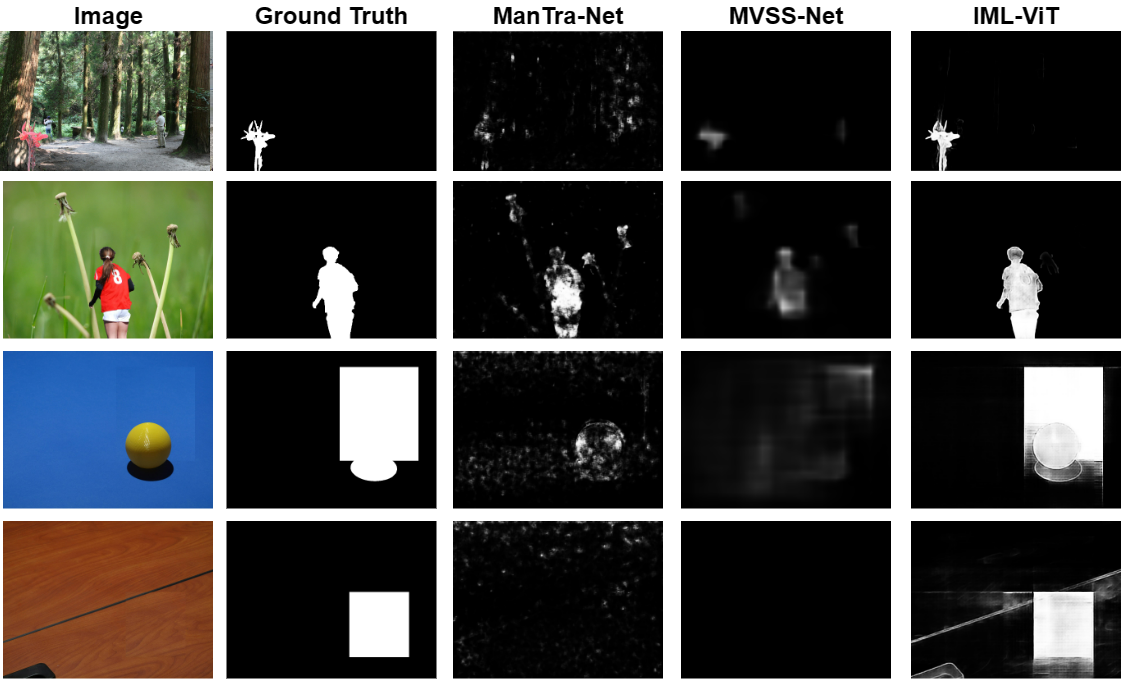}
    \caption{NIST16, some inpainting}
  \end{subfigure}
  \quad
  \begin{subfigure}{0.45\textwidth}
  \centering
    \includegraphics*[width=0.9\columnwidth]{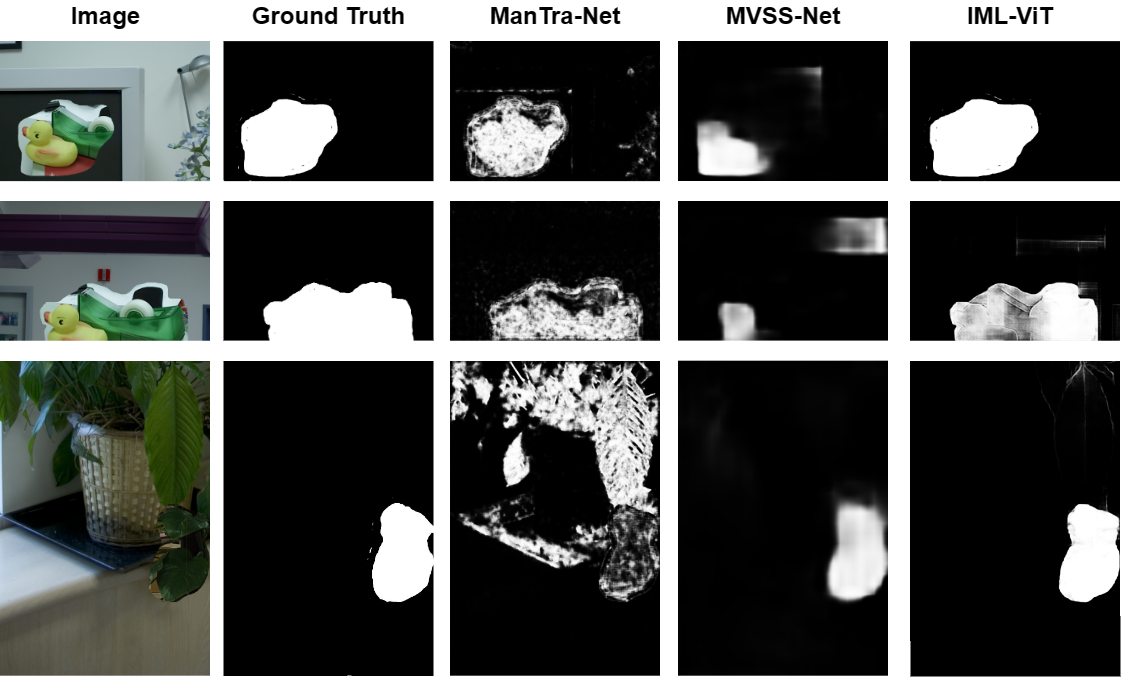}
    \caption{Columbia, all splicing.}
  \end{subfigure}
  \caption{\textbf{Testing results on Protocol No.1 of IML-ViT compare to ManTra-Net and MVSS-Net.} Each dataset has its preference for manipulation types. 
  }
  \label{fig:visualization_app}
\end{figure}

Here we also present some of the predict masks under Protocol No.1, which was from dataset with other preference on manipulation types. Extended from CASIAv1 and COVERAGE datasets in the main paper, we present results in NIST16 and Columbia datasets here in Figure \ref{fig:visualization_app}.

\subsection{Qualitative results for ablation study}
\label{sec:Qualit_ablation}
The ablation study from Figure \ref{fig:ablation} evaluates the impact of various components on IML-ViT's performance: 1) w/o multi-scale: Significant degradation with poor feature detection and blurred outputs. 2) w/o MAE: Improved over the absence of multi-scale, but still blurry with weak edge definition. 3) w/o high-resolution: Noticeable drop in detail and precision, with coarse boundaries. 4) w/o Edge Loss: Less defined edges, preserving overall shape but losing structural details. 5) Full Setup: Produces the most accurate and detailed segmentation maps, capturing fine details and clear boundaries. In summary, the ablation study highlights the critical contributions of each component to the overall performance of IML-ViT. The multi-scale processing, MAE pre-training, high-resolution input, and edge loss each play a vital role in enhancing the model's ability to produce a high-quality segmentation map.

\subsection{Extra results for CASIA datasets.}
To provide a detailed showcase of IML-ViT's performance on image manipulation localization tasks, we present additional image results on the CASIA dataset in Figure \ref{app:casia}.

\begin{figure*}[h]
   \centering
    \includegraphics[width=1.6\columnwidth]{images/app_casia.jpg}
   \caption{\textbf{Additional CASIAv1 results of IML-ViT.} Trained on CASIAv2.}
   \label{app:casia}
\end{figure*}

\subsection{Extra GradCAM results}
\label{sec:extra_gradcam}
\begin{figure*}[h]
   \centering
   \includegraphics[width=2.0\columnwidth]{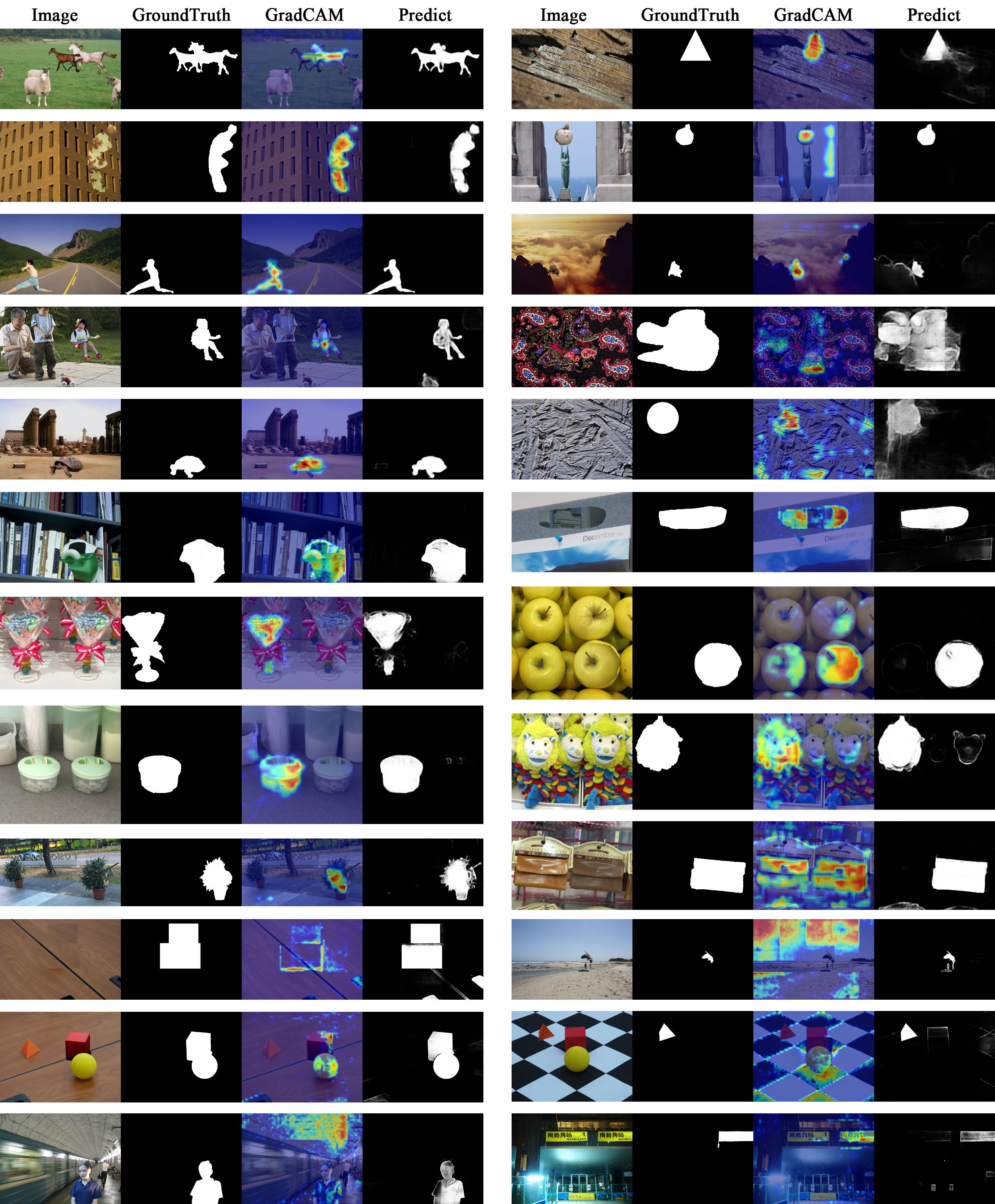}
   \caption{\textbf{Additional GradCAM results.} Datasets are collected from CASIAv1, NIST16, Coverage and Columbia. Attention around the manipulated region and long-range dependency could be observed, which is in line with our motivation to force the model to capture the artifacts and compare the relationships between regions explicitly.}
   \label{app:gradcam}
\end{figure*}

Here we provide extra GradCAM results to verify if IML-ViT focuses on the artifacts we want it to trace. Artifacts are mainly distributed around the manipulated region with rapid changes. Figure \ref{app:gradcam}
vividly shows that the IML-ViT can effectively discover the artifacts from the image and support its decision.

\subsection{Feature maps between each module}
\begin{figure*}[h]
   \centering
   \includegraphics[width=2.0\columnwidth]{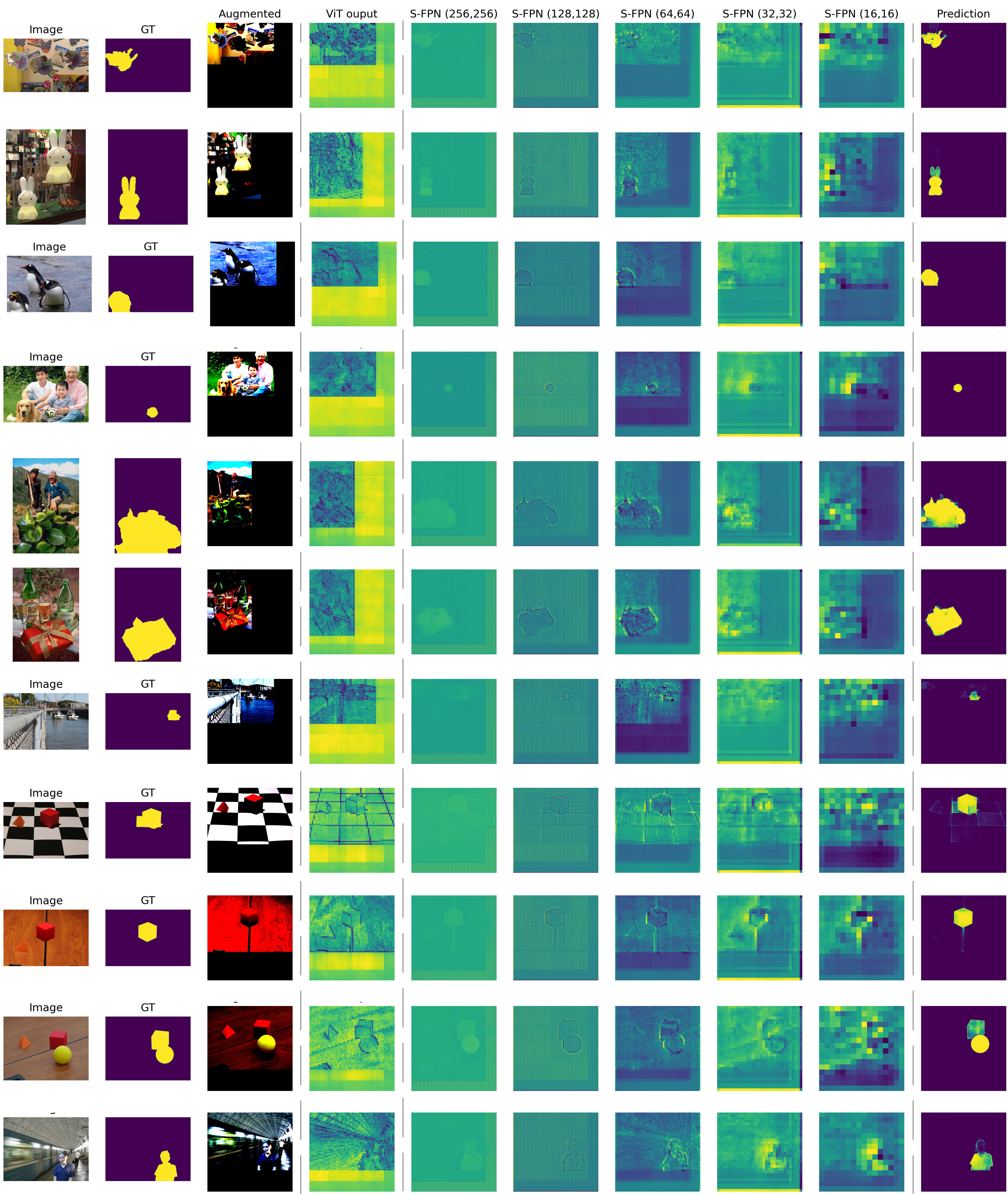}
   \caption{\textbf{Visualization of the outputs from each component.} \textit{GT} denotes ground truth; \textit{Augmented} refers to the image after padding and normalize; \textit{ViT output} is the feature map for the output of ViT backbone; \textit{S-FPN} denotes the respective output of each resolution for different outputs in \textit{simple feature pyramid}. For the output of ViT, since visualization takes the average of all channels, we cannot effectively observe the discrepancies between the manipulated region and the authentic region.  However, we are pleased to see different types of feature expressions in the multi-level output of \textit{S-FPN}. In the high-resolution output, more representation is given to larger, region-level ``contrast differences", while in the (64×64) feature map, we see the image focusing more on edge details and artifacts. This result is in line with the design logic of IML-ViT, which tracks tampering detection from both the perspective of comparing regional low-level differences and capturing detailed visible traces, proving the rationality and effectiveness of our IML-ViT.}
   \label{fig:output}
\end{figure*}

To gain a deeper understanding of IML-ViT, we present visualizations of feature maps between layers by calculating the average channel dimensions of the feature map. The outcomes are displayed in Figure \ref{fig:output}. This visualization process allows us to shed light on the model's functioning and provides valuable insights into its mechanisms.

\section{Limitation}
We observe a rapid decline in IML-ViT's performance on the Gaussian blur attack
when the filter kernel size exceeded 11, We argue that this is mainly because our motivation is to make the model focus on detailed artifacts, but excessive Gaussian blurring can significantly remove these details, leading to a sudden drop in performance. However, from another perspective, this can actually prove that our model is able to effectively capture artifacts in tampering. Currently, the training does not specifically enhance blur, so we believe that adding enough blur data augmentation can compensate for this issue.

\end{document}